\documentclass[acmlarge]{acmart}
\usepackage{svg}
\usepackage{bm}
\usepackage{amsmath}
\usepackage{algorithm}
\usepackage{algpseudocode}
\usepackage{mathtools}

\usepackage{soul}
\usepackage{caption}
\usepackage{subcaption}
\usepackage{multirow}
\usepackage{enumitem}
\usepackage[normalem]{ulem}
\useunder{\uline}{\ul}{}
\usepackage[flushleft]{threeparttable}
\usepackage{setspace}
\usepackage{subcaption}
\usepackage{caption}

\AtBeginDocument{%
\providecommand\BibTeX{{%
		\normalfont B\kern-0.5em{\scshape i\kern-0.25em b}\kern-0.8em\TeX}}}

\newcommand{\etal}{\emph{et al.}}

\newcommand{\eg}{\emph{e.g.},}
\newcommand{\ie}{\emph{i.e.},}

\begin{document}

\title{TAU: Modeling Temporal Consistency Through Temporal Attentive U-Net for PPG Peak Detection}


\author{Chunsheng Zuo}
\authornote{Both authors contributed equally to this research.}
\email{jason.zuo@mail.utoronto.ca}
\orcid{N/A}
\affiliation{%
  \institution{Human Machine Interaction Lab, Huawei Technologies}
  \streetaddress{19 Allstate Pkwy}
  \city{Markham}
  \state{Ontario}
  \country{Canada}
  \postcode{L3R 5A4}
}
\author{Yu Zhao}
\authornotemark[1]
\email{yu.zhao4@huawei.com}
\orcid{N/A}
\affiliation{%
  \institution{Human Machine Interaction Lab, Huawei Technologies}
  \streetaddress{19 Allstate Pkwy}
  \city{Markham}
  \state{Ontario}
  \country{Canada}
  \postcode{L3R 5A4}
}
\author{Juntao Ye}
\email{juntao.ye@huawei.com}
\orcid{N/A}
\affiliation{%
  \institution{Human Machine Interaction Lab, Huawei Technologies}
  \streetaddress{19 Allstate Pkwy}
  \city{Markham}
  \state{Ontario}
  \country{Canada}
  \postcode{L3R 5A4}
}

\begin{abstract}
Photoplethysmography (PPG) sensors have been widely used in consumer wearable devices to monitor heart rates (HR) and heart rate variability (HRV). Despite the prevalence, PPG signals can be contaminated by motion artifacts induced from daily activities. Existing approaches mainly use the amplitude information to perform PPG peak detection. However, these approaches cannot accurately identify peaks, since motion artifacts may bring random and significant amplitude variations. To improve the performance of PPG peak detection, the time information can be used. Specifically, heart rates exhibit temporal consistency that consecutive heartbeat intervals in a normal person can have limited variations. To leverage the temporal consistency, we propose the Temporal Attentive U-Net, \ie{} TAU, to accurately detect peaks from PPG signals.  In TAU, we design a time module that encodes temporal consistency in temporal embeddings. We integrate the amplitude information with temporal embeddings using the attention mechanism to estimate peak labels. Our experimental results show that TAU outperforms eleven baselines on heart rate estimation by more than 22.4\%. Our TAU model achieves the best performance across various Signal-to-Noise Ratio (SNR) levels. Moreover, we achieve Pearson correlation coefficients higher than 0.9 ($p$ < 0.01) on estimating HRV features from low-noise-level PPG signals.
\end{abstract}

%



\maketitle
\section{Introduction} 
With the prevalence of consumer-grade smart wearable devices, cardiac monitoring is accessible in our daily lives.
Cardiac monitoring continuously or intermittently detects the intervals between heartbeats (\ie{} cardiac cycles) to derive heart rates (HR) and heart rate variability (HRV).
HR and HRV can be served as early signs to predict the risk of cardiovascular disease~\cite{bayoumy2021smart}\cite{kubota2017heart}\cite{palatini1997heart}. 
For example, a high heart rate at rest in healthy populations may indicate a high risk of coronary artery disease~\cite{zhang2016association}.
Irregular heartbeat intervals may lead to cardiac arrhythmia~\cite{fenton2008cardiac} and indicate heart failure~\cite{bayoumy2021smart}.

\begin{figure}[t]
	\centering
	\includegraphics[width=0.4\textwidth]{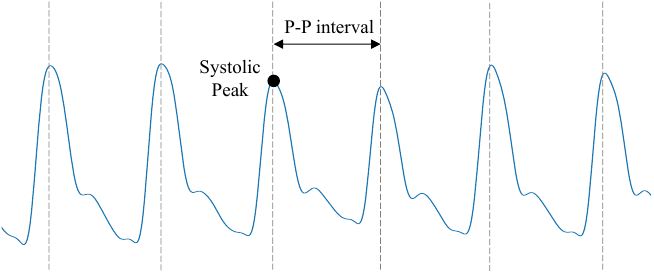}
	\caption{An illustration of PPG signals, systolic peaks and P-P intervals.}
	\label{fig:PPG}
	\vspace{-10pt}
\end{figure}

Two commonly used signals in cardiac monitoring are Electrocardiography (\ie{} ECG) and Photoplethysmography (\ie{} PPG).
In particular, ECG records electrical changes from cardiac cycles using electrodes placed on the skin.
Due to the high accuracy in measuring heartbeats, ECG has become a standard diagnostic tool to monitor patient status in clinical settings. 
However, few portable wearable devices can record ECG signals, limiting the wide adoption among consumers~\cite{bayoumy2021smart}.
PPG sensors are optical sensors that emit and detect light to measure the changes of blood volumes in blood vessels~\cite{castaneda2018review}.
Since PPG sensors are inexpensive and portable, PPG sensors are widely adopted in commercial wearable devices to detect heart rates and heart rate variability.
Depending on the positions of light-emitters and photodetectors, PPG sensors can have transmission mode and reflectance mode~\cite{castaneda2018review}.
Transmission mode PPG sensors are placed on fingertip and earlobe, where light-emitters and photodetectors are separated by the skin.
In the reflectance mode, photodetectors and light-emitters are positioned on the same side of the skin to measure the reflected light.
Reflectance mode PPG sensors are commonly used in commercial smartwatches to record PPG signals.
A PPG signal corresponds to two phases in a cardiac cycle, \ie{} systolic phase and diastolic phase~\cite{castaneda2018review}\cite{elgendi2012analysis}. 
At the systolic phase, a PPG signal reaches the systolic peak (shown in Figure~\ref{fig:PPG}) when the blood volume that is pumped by a heartbeat reaches a local maximum.
At the diastolic phase, the heart muscle relaxes and a PPG signal reaches the valley.
The interval between two systolic peaks is the Peak-to-Peak interval (\ie{} P-P interval) that quantifies the heartbeat interval.
Heartbeat intervals exhibit temporal consistency that two heartbeat intervals at a short time scale cannot vary by a large amount~\cite{avram2019real}\cite{de2020ppg}\cite{salehizadeh2015SpaMa}.
Salehizadeh~\etal~\cite{salehizadeh2015SpaMa} suggest that two consecutive heartbeat intervals should have a variation less than 10 beats per minute (\ie{} 54ms to 200ms depending on heart rates).

Despite the prevalence in HR and HRV detection, PPG signals can be contaminated by motion artifacts (MA)~\cite{afandizadeh2023accurate}\cite{kim2006motion}\cite{Sarkar_Etemad_2021}.
Daily routine activities, such as hand movement, may loosen the contact between PPG sensors in smartwatches and the skin, causing ambient light to be absorbed by PPG sensors~\cite{patterson2011ratiometric}.
Therefore, the noise in PPG signals presents a significant challenge to accurately localize systolic peaks and estimate heartbeat intervals. 
Research has been invested considerably to detect peaks from PPG signals.
Generally, existing approaches can be classified into two categories: (1) signal-processing approaches and (2) deep learning approaches.
Many \textit{signal-processing approaches}~\cite{chakraborty2020hilbert}\cite{elgendi_norton_brearley_abbott_schuurmans_2013}\cite{kuntamalla2014adaptive}\cite{VANGENT2019368} determine peak positions based on the amplitude information to find local maxima.
However, the approaches cannot accurately identify peaks in noise contaminated  PPG signals, since motion artifacts may bring random and significant amplitude variations.
\textit{Deep learning approaches}~\cite{Kazemi20226054}\cite{Sarkar_Etemad_2021} observe peaks in noisy PPG signals from the training set and apply deep learning networks, \eg{} convolutional neural networks (CNNs), to infer peak positions in unseen PPG signals.
The approaches may not generalize well to PPG signals whose noise patterns are not observed in the training set.
When amplitude values are distorted by noise, the time information can be used to determine peak positions in PPG signals.
Specifically, P-P intervals in a PPG signal should comply with the temporal consistency of heartbeat intervals.
However, none of the above-mentioned approaches explicitly model the time information to detect peaks.
Other deep learning approaches~\cite{8607019cornet}\cite{Reiss20193079}\cite{8856989PPGNET}\cite{s21155212MHCAPPG} directly estimate heart rates from PPG signals without performing peak detection.
The approaches cannot extract heart rate variability.

To accurately detect peaks from noisy PPG signals, we propose a deep learning model, \ie{} Temporal Attentive U-Net (TAU), that explicitly models the time information.
Similar to U-Net~\cite{ronneberger2015u}, our TAU model adopts an encoder-decoder architecture.
The encoder module in TAU aggregates the local context of PPG signals to generate vectorized representations (\ie{} embeddings) of the amplitude information.
To augment amplitude embeddings, we design a time module that encodes the time information to temporal embeddings.
Specifically, for each PPG sample in a PPG signal, we derive a \textit{sample-to-peak} distance sequence that computes the distance between the sample with nearby peaks.
We compare each value in the distance sequence with cardiac cycles of PPG signals using the attention mechanism~\cite{vaswani2017attention}, to determine whether the value is temporally consistent with cardiac cycles.
The values in a \textit{sample-to-peak} distance sequence are mapped to temporal embeddings and aggregated to represent the time information of a PPG sample.
The decoder module integrates temporal embeddings and amplitude embeddings using the attention mechanism to capture the global context of PPG signals.
To obtain ground truth labels for supervised learning, we use distance transform~\cite{ma2020distance} to derive the label of a PPG sample as the distance between the sample and a nearby peak.
Distance transform labels ensure that semantic similar PPG samples have similar label values.
In summary, we make the following contributions:
\begin{itemize}[wide=5pt] 
	\item We design a time module that models the temporal consistency of PPG signals using temporal embeddings.
	Our model generalizes to unseen PPG noise patterns to detect peaks, by leveraging the common properties of PPG signals that P-P intervals should have limited variations.
	
	\item We propose an encoder-decoder architecture to localize peaks in PPG signals.
	The encoder module uses convolution operations to capture the local context of the amplitude information.
    The decoder module employs the attention mechanism to aggregate the global context of the amplitude information and the time information to estimate peak labels.
	The estimated peak labels are supervised by distance transform labels to ensure that semantic similar PPG samples get similar labels.
	To promote the usage of our model on resource constraint devices, we develop a lightweight model TAU-lite.
	
	\item We conduct extensive experiments to compare the performance of our model with eleven state-of-the-art baselines on four public datasets, \ie{} BIDMC, CAPNO, WESAD, and DALIA.
    The four datasets are collected from both transmission mode PPG sensors and reflectance mode PPG sensors.
	We perform three tasks: (1) peak detection; (2) heart rate estimation; and (3) HRV estimation.
	Our results show that, on peak detection, TAU outperforms the best performing baseline 1D-CNN-DT~\cite{Kazemi20226054} by 2.8\% on F1-score.
        On heart rate estimation, we perform experiments using datasets with two window sizes, \ie{} 10-second and 8-second.
	Our designed time module improves the performance of heart rate estimation, as TAU outperforms the eleven baselines by more than 22.4\%.
        We compute Signal-to-Noise Ratio (SNR) of PPG signals and observe that our TAU model achieves the best heart rate estimation performance in both low-noise-level signals (SNR > -4.6dB) and high-noise-level signals (SNR < -9.1dB).
        On HRV estimation, our TAU and TAU-lite models obtain Pearson correlation coefficients higher than 0.9 ($p$ < 0.01) when estimating HRV features using the BIDMC dataset.
	Moreover, TAU-lite significantly reduces the inference time of TAU and achieves similar inference time to 1D-CNN that uses convolution operations.
	
\end{itemize}

\vspace{5pt}
\noindent \textbf{Paper Organization.}
Section~\ref{sec:related_work} describes the related work.
Section~\ref{sec:approach} presents our proposed TAU model.
Section~\ref{sec:setup} discusses the experimental setup.
Section~\ref{sec:results} illustrates our obtained results.
Finally, Section~\ref{sec:conclusion} concludes our paper.

\section{Related Work}\label{sec:related_work}

In this section, we summarize the related work on time-domain signal-processing approaches, frequency-domain signal-processing approaches, deep learning approaches for signal processing, heart rate variability (HRV) analysis and temporal modeling in deep learning.

\vspace{5pt}
\subsection{Time-Domain Signal-Processing Approaches}
Time-domain signal-processing approaches identify peaks from PPG signals using the amplitude information.
Existing approaches~\cite{chakraborty2020hilbert}\cite{elgendi_norton_brearley_abbott_schuurmans_2013}\cite{kuntamalla2014adaptive}\cite{VANGENT2019368} generally take three steps to perform peak detection: 1) preprocessing signals to remove noises. Specifically, Vadrevu~\etal~\cite{vadrevu2018robust} use stationary wavelet transform~\cite{nason1995stationary} to remove high-frequency and low-frequency noises from PPG signals.
Alqaraawi~\etal~\cite{alqaraawi2016bayesampd} build a probabilistic model that computes the probability of a sample to be a peak given historical peak information;
(2) identifying regions of interest that may contain potential peaks.
For example, adaptive threshold based approaches~\cite{kuntamalla2014adaptive}\cite{van2019analysing} update a threshold adaptively according to amplitude values of PPG signals. 
The adapted thresholds are used to select regions that may contain peaks.
and (3) determining peaks by finding local maxima on regions of interest.
The obtained peak-to-peak intervals can be used to estimate heart rates and heart rate variability.
Time-domain signal-processing approaches may not be able to filter out motion artifact contaminated noise in PPG signals~\cite{Kazemi20226054}.


\subsection{Frequency-Domain Signal-Processing Approaches}

Frequency-domain signal-processing approaches~\cite{al2023adaptive} estimate heart rates by extracting the frequency spectrum of a PPG segment.
Generally, the approaches mainly consist of three steps~\cite{biswas2019heart}:
(1) motion artifact removal using noise subtraction and blind source separation (BSS)~\cite{salehizadeh2015SpaMa} approaches.
Noise subtraction approaches~\cite{al2023adaptive}\cite{fujita2017parhelia}\cite{lee2018wearable}~\cite{zhang2014troika} usually use accelerometer signals as reference signals to remove motion artifacts from PPG signals.
For example, some approaches~\cite{al2023adaptive}\cite{arunkumar2020robust}\cite{jarchi2017towards}\cite{wen2023wearable}\cite{ye2016combining} apply adaptive filters on accelerometer signals to denoise PPG signals.
TROIKA~\cite{zhang2014troika}, JOSS~\cite{zhang2015photoplethysmography} and SpaMA~\cite{salehizadeh2015SpaMa} perform frequency analysis to remove motion artifact induced frequencies that are present in both the PPG spectrum and the accelerometer spectrum. 
BSS approaches aim to decompose noisy PPG signals to extract clean signals without using accelerometers.
Common approaches to decompose noisy PPG signals are independent component analysis (ICA)~\cite{krishnan2010FDICA}, principle component analysis~\cite{motin2017ensemble}, singular spectrum analysis~\cite{salehizadeh2014photoplethysmograph} and ensemble empirical mode decomposition~\cite{garde2013empirical}\cite{khan2015EEMD};
(2) heart rate estimation by analyzing the spectra of PPG signals.
Specifically, Wen~\etal~\cite{wen2023wearable} build decision tree models to predict frequency searching intervals to identify frequencies with maximum spectral values; and 
(3) heart rate correction based on the assumption that there are no significant changes between two consecutive heart rate estimations~\cite{guo2022effective}.
In particular, WFPV~\cite{temko2017WFPD} uses a phase vocoder~\cite{flanagan1966phase} to improve the resolution of the frequency spectrum and refine the dominant frequency value.
Koneshloo and Du~\cite{koneshloo2019novel} use the Lasso model to reconstruct the spectra of PPG signals using the spectra in previous windows.
SPECTRAP~\cite{sun2015SPECTRAP} models the probability that a frequency bin on a spectrum represents the heart rate.
All the above-mentioned approaches obtain average heart rates from PPG signals and cannot measure HRV.

\subsection{Deep Learning Approaches}
Several deep learning approaches use feed-forward neural networks~\cite{ahmed2023DLWL}, CNNs~\cite{chang2021deepheart}, autoencoders~\cite{mohagheghian2023noise} and CycleGANs~\cite{afandizadeh2023accurate}, to reconstruct clean PPG signals from noisy PPG signals.
Approaches such as PPG2ECGps~\cite{tang2023ppg2ecgps} and CardioGAN~\cite{Sarkar_Etemad_2021} focus on denoising PPG signals to generate ECG signals.
Since peaks from the generated signals may not be aligned with peaks in the corresponding PPG signals, these approaches detect peaks from reconstructed signals to estimate heart rates.
Similar to our TAU model, 1D-CNN~\cite{Kazemi20226054} predicts peaks from PPG signals using dilated convolutions.
Approaches such as Yen~\etal~\cite{yen2021applying} and Panwar~\etal~\cite{panwar2020pp} adopt CNN and long short-term memory (LSTM) to estimate multiple physiological features simultaneously from PPG signals, \eg{} blood pressure and heart rates.
Other approaches~\cite{8607019cornet}\cite{Reiss20193079}\cite{8856989PPGNET}\cite{s21155212MHCAPPG} directly predict heart rates.
For example, DeepPPG~\cite{Reiss20193079} uses the time-frequency spectra of PPG and accelerometer signals as the inputs to a CNN based model.
Wilkosz and Szczęsna~\cite{s21155212MHCAPPG} use PPG and accelerometer signals from the time domain as inputs.
CorNet~\cite{8607019cornet} and PPGnet~\cite{8856989PPGNET} build CNN and LSTM architectures to predict heart rates from PPG signals in the time domain.

\subsection{Heart Rate Variability Analysis}
Estimating HRV from PPG signals is more challenging compared to heart rate estimations, since both motion artifacts and HRV are inconsistent in consecutive time windows~\cite{aygun2019robust}.
Aygun~\etal~\cite{aygun2019robust} identify systolic peaks, maximum slopes and onsets of PPG signals as candidate nodes. Candidate nodes are connected using a shortest path algorithm to estimate heartbeat intervals.
Jarchi~\etal~\cite{jarchi2017towards} apply adaptive filtering and Hilbert transform to obtain time-domain HRV features.
Shuffle-rPPGNet~\cite{kuang2023shuffle} builds a 3-D CNN model to estimate HRV features from facial videos.
Georgiou~\etal~\cite{georgiou2018can} report that HRV cannot be estimated from PPG signals with high levels of motion artifacts.
Because of motion artifacts, limited frequency range and imprecise peak detection in PPG signals, estimating HRV from PPG signals is less accurate than estimating HRV from ECG signals~\cite{georgiou2018can}\cite{jarchi2017towards}\cite{umair2021hrv}. 
Umair~\etal~\cite{umair2021hrv} find that the accuracy of HRV estimations may vary among different types of measurement devices.
In clinical settings, the optimal sampling rate to estimate HRV from ECG signals is 250 to 500Hz~\cite{electrophysiology1996heart}.
Kwon~\etal~\cite{kwon2018electrocardiogram} reveal that ECG signals sampled at 100Hz can extract accurate time-domain HRV features.

\subsection{Temporal Modeling in Deep Learning}
Many deep learning approaches explicitly model the time information in various application domains, \eg{} natural language understanding.
Specifically, Time2Vec~\cite{kazemi2019time2vec} represents time as a learnable vector that captures periodic and non-periodic behaviors of events.
Goyal and Durrett~\cite{goyal2019timeexpress} encode time expressions in sentences with embeddings.
The learned temporal embeddings capture temporal orderings of events. 
Zhou~\etal~\cite{zhou2022tatransrec} use the time interval between two consecutive online courses to perform online course recommendations. 
Informer~\cite{zhou2021informer} adds timestamps to the representations of time-series data to facilitate long sequence time-series forecasting.
Wu~\etal~\cite{wu2020DAtrans} propose a distance-aware transformer architecture that uses the distances between tokens in a sentence to rescale attention weights.
Zhang~\etal~\cite{zhang2020tatranscli} design a time-aware layer that captures temporal importance of clinical notes to classify a patient's health states.
Similar to the above-mentioned approaches, our TAU model explicitly models the time information.
We model the temporal consistency of heartbeat intervals using temporal embeddings to improve the performance of peak detection in noisy PPG signals.

\section{The Proposed Temporal Attentive U-Net Model}\label{sec:approach}
Our TAU model aims to identify positions of systolic peaks from segments of PPG signals (\ie{} PPG segments) that are captured by wearable devices (\eg{} smartwatches) in real-world settings. 
In this section, we present the details of our proposed TAU model.
First, we describe our approach to generate ground truth peak labels to supervise model training.
Then, we discuss the model architecture of TAU.

\subsection{Labeling Peaks with Distance Transform}\label{sec:dt}
\begin{figure}[htbp]
	\centering
    \includegraphics[width=0.5\textwidth]{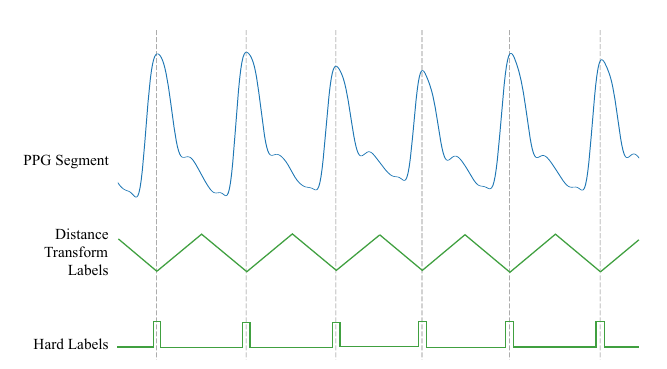}
    \vspace{-5pt}
	\caption{An illustration of distance transform labels.}
	\label{fig:Distance Transform}
 \vspace{-10pt}
\end{figure}

In a PPG segment, the number of systolic peaks are much less compared to the number of PPG samples.
Therefore, labeling the position of systolic peaks as 1 and non-systolic peaks as 0 may cause the label sparsity issue and significantly degrade model performance~\cite{yang2023self}\cite{zhao2022modeling}.
Kianoosh~\etal~\cite{Kazemi20226054} use a hard labeling schema that labels a number of PPG samples (\eg{} five PPG samples) around the peak position as 1 and the rest as 0.
However, the number of peaks in hard labels is still much less compared to the number of non-peaks (shown in Figure~\ref{fig:Distance Transform}).
Moreover, closely positioned PPG samples may have semantically similar representations due to the similar temporal order and amplitude values.
In hard labels, the sharp transitions between label classes (\ie{} 0 and 1)  may lose the continuity and convexity of PPG signals.

To address the label sparsity issue and persist the continuity of PPG signals, we propose the soft labeling schema that labels peaks with distance transform~\cite{audebert2019distance}\cite{ma2020distance}.
Motivated by the computer vision community, we use the distance transform to calculate the distance between the position of a PPG sample and the position of the peak that is closest to the PPG sample (shown in Figure~\ref{fig:Distance Transform}).
To be specific, let $P$ be a set that contains positions of peaks in a PPG segment $\bm{s}$. 
The label $y_i$ of a PPG sample at the position $i \in [1,n]$ ($n$ denotes the number of PPG samples) is calculated as:
\begin{equation}\label{eq:dt}
	y_i = \min_j | {i-P_j}|,
\end{equation}

\noindent where $P_j$ represents the position of the $j$th peak in the set $P$.

PPG segments could be sampled at different sampling frequencies to represent the same cardiac cycles.
To generate labels for a PPG segment $\bm{s'}$ with a number $n'$ of PPG samples, we use Equation~\ref{eq:dt_scale}:
\begin{equation}\label{eq:dt_scale}
	y'_i = \lfloor \frac{n}{n'} \times \min_j({i-P'_j}) \rfloor,
\end{equation}
\noindent where $P'$ denotes the set of peak positions of a PPG segment $\bm{s'}$, $i \in [1,n']$ is the position of a PPG sample in a PPG segment $\bm{s'}$, and the operator $\lfloor \rfloor$ rounds the value to the nearest integer.
The multiplier $\frac{n}{n'}$ in Equation~\ref{eq:dt_scale} ensures that labels for PPG segments $\bm{s}$ and $\bm{s'}$ with different sampling frequencies are on the same scale.

With our soft labeling schema, the labels for systolic peaks are 0.
The larger the label value, the less likely the PPG sample is a peak.
To identify positions of systolic peaks from label predictions, we devise a peak searching function.
The peak searching function identifies regions of interest with consecutive predicted labels that are below a threshold value. 
The position of the minimum label value within a region of interest is identified as the position of a peak.

\subsection{Model Architecture}\label{sec:arch}
\begin{figure}[t]
	\centering
    \includegraphics[width=0.99\textwidth]{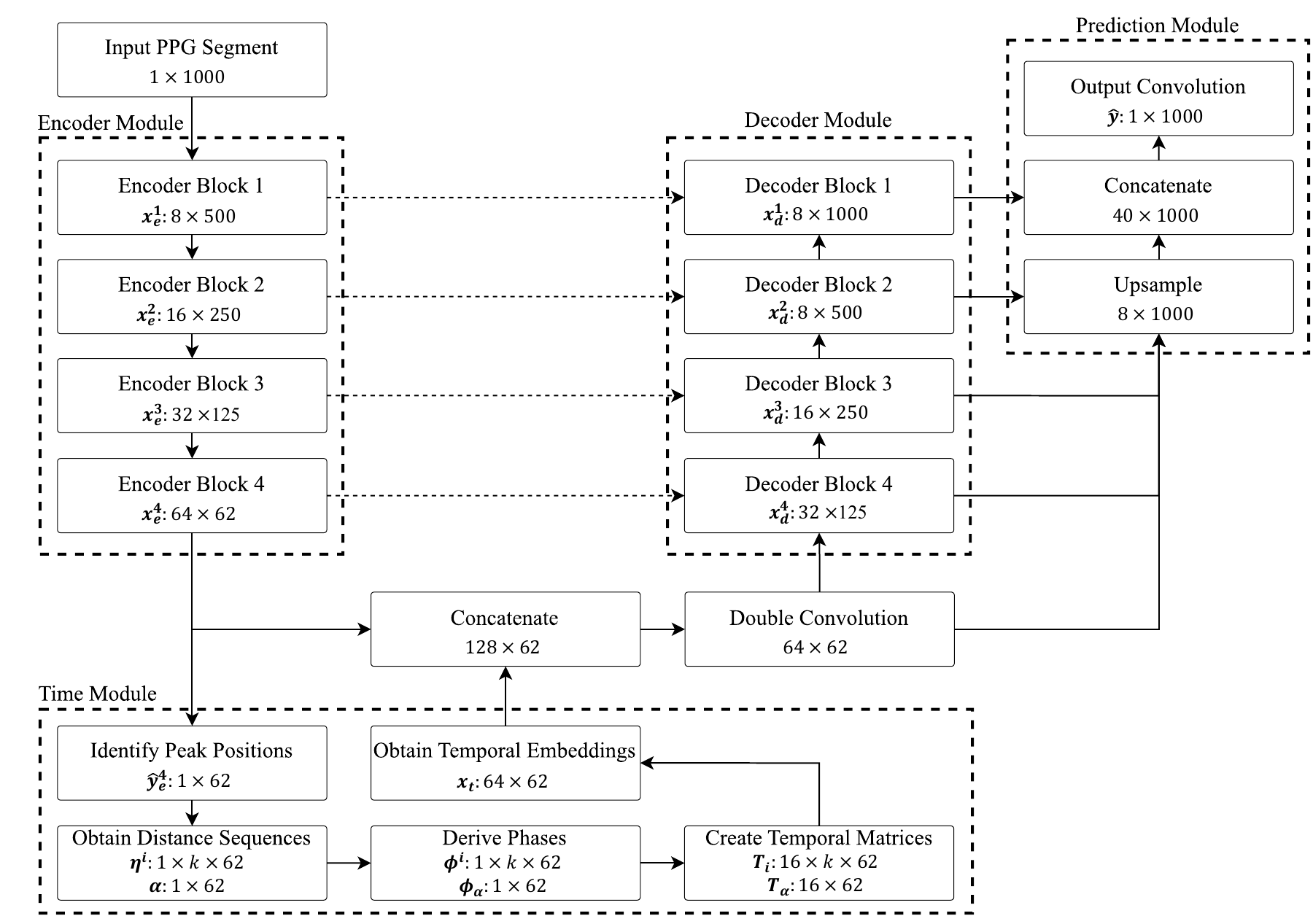}
	\caption{The architecture of our proposed TAU model.}
	\label{fig:architecture}
\end{figure}

TAU employs an architecture similar to U-Net~\cite{ronneberger2015u}, which consists of four modules: an encoder module, a time module, a skip-connected decoder module, and a prediction module.
Figure~\ref{fig:architecture} shows the architecture of TAU.
The encoder module aggregates PPG samples to learn the amplitude information.
Since the amplitude information may be ambiguous to identify peaks, the time module models the time information to compare the distance between PPG samples.
The decoder module combines the amplitude information and the time information to generate peak label predictions.
Finally, the prediction module supervises the training of peak label predictions using distance transform labels.
In the following, we describe the four modules in detail.

\subsubsection{Encoder Module}\label{sec:encoder}
The encoder module consists of a number $L$ of encoder blocks to capture the local context of PPG amplitude values.
An encoder block contains a double convolution layer and a max-pooling layer.
A double convolution layer performs two 1D dilated convolution operations to double the number of feature channels (\ie{} embedding size)~\cite{Kazemi20226054}.
In the first encoder block, the double convolution layer augments the number of feature channels $C_0$ to 8. 
A max-pooling layer performs a 1D max-pooling operation with a pool-size of 2 to downsample the number of PPG samples.
After a number $L$ of encoder blocks, we apply a double convolution layer to generate PPG embeddings $\bm{x_e^L}$.

\subsubsection{Time Module}\label{sec:time}
A PPG sample is likely to be a peak, if the distances between a PPG sample and nearby peaks are temporally consistent with cardiac cycles of the PPG signal.
The time module explicitly models the distance using temporal embeddings. 
We build temporal embeddings from the output $\bm{x_e^L}$ of the encoder module using the following five steps.

\begin{figure}[t]
	\centering
    \includegraphics[width=0.5\textwidth]{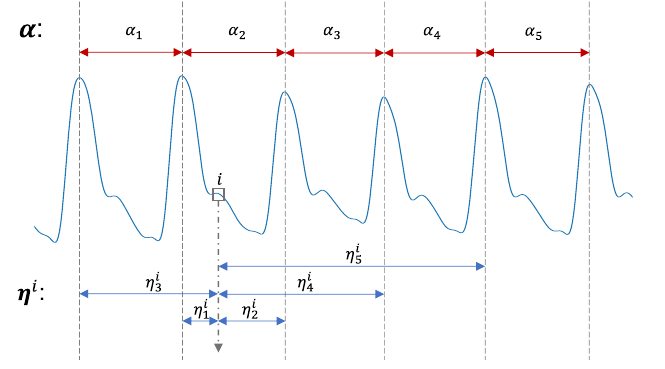}
	\caption{An illustration of \textit{peak-to-peak} distance sequence and \textit{sample-to-peak} distance sequence.}
	\label{fig:distance_sequence}
 \vspace{-15pt}
\end{figure}

\textit{1) Identify Peak Positions.}
To identify distances between PPG samples and peaks, we predict peak labels $\bm{\hat{y}_e^L}$ from the output $\bm{x_e^L}$ using Equation~\ref{eq:regression}:
\begin{equation}\label{eq:regression}
    \bm{\hat{y}_e^L} = \sigma(\bm{W_e}\bm{x_e^L} + b_e),
\end{equation}
\noindent where $\bm{W_e}$ is a weight matrix, $\bm{b_e}$ is the bias, and $\sigma$ is the non-linear activation function, \ie{} ReLU.
We supervise the learning of label predictions $\bm{\hat{y}_e^L}$ using distance transform labels $\bm{y_e^L}$ that are generated from Equation~\ref{eq:dt_scale}.
Then, we extract peak positions from label predictions $\bm{\hat{y}_e^L}$ using the peak searching function as described in Section~\ref{sec:dt}.

\textit{2) Obtain Distance Sequences.}
With the identified peak positions, we obtain two sequences of distance values:
(1) a \textit{peak-to-peak} distance sequence $\bm{\alpha}$ (shown in Figure~\ref{fig:distance_sequence}), which is formulated by calculating the distance between each pair of neighbor peaks.
We choose the median value $m_\alpha$ in $\bm{\alpha}$ as the reference distance to represent the cardiac cycle of the PPG segment.
The median value is not skewed by extremely large or small distance values to represent the center of the peak-to-peak distance sequence; and
(2) a \textit{sample-to-peak} distance sequence $\bm{\eta}^i$ (shown in Figure~\ref{fig:distance_sequence}) for a PPG sample $i$, which is formulated by calculating the distance between the PPG sample $i$ and its $k$-closest peaks.

\textit{3) Derive Phases.}
A PPG sample is likely to be a peak if the PPG sample completes full cardiac cycles to reach other peak positions.
To determine whether distance values complete full cardiac cycles, we derive phases.
In our approach, a phase $\phi_j^i$ between a PPG sample $i$ and a peak $j$ denotes the minimum distance that the PPG sample $i$ needs to travel to complete full cardiac cycles with the peak $j$.
Given the distance $\eta_j^i$ between a PPG sample $i$ and a peak $j$, we use Equation~\ref{eq:phase} to calculate the phase $\phi_j^i$:
\begin{equation}\label{eq:phase}
	\begin{aligned}
		r_j^i &= \eta_j^i \bmod m_\alpha, \\
		\phi_j^i &= \min(m_\alpha - r_j^i,  r_j^i),
	\end{aligned}
\end{equation}

\noindent where $mod$ denotes the modulo operation, $r_j^i$ is the remainder after taking the modulo operation, and $\phi_j^i \in [0, \frac{m_\alpha}{2}]$. 
We compute phases for each distance value in distance sequences $\bm{\alpha}$ and $\bm{\eta}^i$.
A phase value $\phi_j^i$ of 0 represents that the PPG sample $i$ completes full cardiac cycles to reach the peak $j$.
A phase value reaches the maximum value when the remainder $r_j^i$ is at the middle of the cardiac cycle (\ie{} $r_j^i = \frac{m_\alpha}{2}$).
Intuitively, the larger the phase value, the less likely the PPG sample is a peak.

\textit{4) Create Temporal Matrices.} 
To incorporate more temporal information, we project each derived phase value to an embedding vector $\bm{e} \in \mathcal{R}^E$, where $E$ is the embedding size ($E$=16 in Figure~\ref{fig:architecture}). 
The embedding vectors $\bm{e}$ are randomly initialized with a Gaussian distribution $\mathcal{N}(0,1)$ and optimized during the training process.
Since PPG signals reflect cardiac cycles, the order of PPG samples in a PPG segment cannot be altered.
To model the position information of a PPG sample, we generate a positional embedding vector $\bm{p}$ for a PPG sample using the sinusoidal functions, similar to the work by Vaswani~\etal~\cite{vaswani2017attention}.
Adding the embedding $\bm{e}$ and the positional embedding $\bm{p}$, we derive the temporal embedding vector $\bm{\tau} = \bm{e} + \bm{p}$ for a distance value in distance sequences $\bm{\alpha}$ and $\bm{\eta}^i$.
Packing temporal embeddings into matrices, we derive a temporal matrix $\bm{T_\alpha}$ for the distance sequence $\bm{\alpha}$, and a temporal matrix $\bm{T_{i}}$ for the distance sequence $\bm{\eta^i}$.

\textit{5) Obtain Temporal Embeddings.}
To obtain a temporal embedding vector for a PPG sample $i$, we compare the temporal information in the distance sequence $\bm{\eta^i}$ with the temporal information of peaks in the distance sequence $\bm{\alpha}$.
Specifically, we implement the multi-head attention mechanism, using the temporal matrix $\bm{T_{i}}$ as the query matrix, and the temporal matrix $\bm{T_\alpha}$ as the key and value matrices, respectively.
The query matrix $\bm{T_{i}}$ is residually added to the output $\bm{O_{i}}$ from the multi-head attention mechanism to derive the temporal matrix $\bm{U_{i}}$, \ie{} $\bm{U_{i}} = \bm{O_{i}} + \bm{T_{i}}$.
The residual connection accelerates model training.
We take the average of $k$ temporal embeddings in $\bm{U_{i}}$ to obtain a temporal embedding vector $\bm{u_i} \in \mathcal{R}^E$ of the PPG sample $i$.
To associate the semantics of temporal embeddings with peaks, we use Equation~\ref{eq:regression} to predict peak labels $\bm{\hat{y}_t}$ from temporal embeddings $\bm{u_i}$ of PPG samples.
The learning of peak label predictions $\bm{\hat{y}_t}$ is supervised by distance transform labels $\bm{y_e^L}$.
Temporal embeddings $\bm{u_i}$ are applied with a double convolution layer to generate temporal embeddings $\bm{x_{t}}$ that has the same dimension as the output $\bm{x_{e}^{L}}$ of the encoder module.

\subsubsection{Decoder Module}\label{sec:decoder}
We concatenate sample embeddings $\bm{x_{e}^{L}}$ and temporal embeddings $\bm{x_{t}}$ and apply a double convolutional layer to generate the input of the decoder module.
The decoder module consists of a number of decoder blocks.
The $l^{th}$ decoder block sequentially applies a multi-head attention layer, an upsample layer, and a double convolution layer to generate PPG embeddings $\bm{x_d^l}$. 
A multi-head attention layer learns long-term dependencies among PPG samples and aggregates embeddings of PPG signals globally using the multi-head attention mechanism.
An upsample layer first applies two 1D dilated convolution operations, and then linearly upsamples the number of PPG samples by a factor of 2. 
The double convolution layer reduces the number of feature channels to keep the same dimensions as the input of the $l^{th}$ encoder block.
A decoder module and an encoder module are skip connected to facilitate model convergence~\cite{ronneberger2015u}.

\subsubsection{Prediction Module}\label{sec:prediction}
After a number $L$ of decoder blocks, the prediction module predicts peak labels.
Specifically, we apply an upsample layer on PPG embeddings $\bm{x_d^l}$ from each decoder block to keep the  dimensions of upsampled PPG embeddings the same.
Motivated by residual connections~\cite{cai2022ma}\cite{he2016identity}, we concatenate the upsampled PPG embeddings of each decoder block and apply an output convolution layer to generate peak labels $\bm{\hat{y}}$.
An output convolution layer consists of three 1D convolution operations.
To further enhance feature learning in decoder blocks, we adopt the deep supervision strategy that is proposed by Lee~\etal~\cite{lee2015deeply} to estimate peak labels at each decoder block.

We optimize the estimated label predictions $\bm{\hat{y}}$ by minimizing the smooth $L1$ loss function~\cite{girshick2015fast}:
\begin{equation}\label{eq:smooth_loss}
	L1_{smooth} (\bm{\hat{y}}, \bm{y}) = 
	\begin{cases}
		\frac{1}{2}(\bm{\hat{y}}-\bm{y})^2, & \text{if} \; |\bm{\hat{y}}-\bm{y}|<1  \\
		|\bm{\hat{y}}-\bm{y}| - \frac{1}{2}, & \text{otherwise}, \\
	\end{cases}
\end{equation}  
\noindent where $\bm{y}$ is the ground truth that is labeled with distance transform (see Section~\ref{sec:dt}).
The smooth $L1$ loss function is a combination of the squared error loss (\ie{} $L2$ loss) and absolute error loss (\ie{} $L1$ loss).
When predictions and ground truth have small absolute errors (\ie{} $|\bm{\hat{y}}-\bm{y}|<1$), the $L2$ loss allows small gradient values to slow down model training.
For large absolute errors, the $L1$ loss provides a constant gradient to
stabilize the training process.

\section{Experimental Setup}\label{sec:setup}
In this section, we describe our datasets, evaluation metrics, hyperparameter selection, our lightweight model, \ie{} TAU-lite, and baselines.

\subsection{Datasets}\label{sec:dataset}

\textbf{Datasets.} 
We perform experiments on four public available datasets, \ie{} BIDMC, CAPNO, WESAD, and DALIA, similar to the work by Sarkar and Etemad~\cite{Sarkar_Etemad_2021}.
\begin{itemize}[]
	\item \textit{BIDMC}~\cite{Pimentel7748483BIDMC} contains ECG and PPG signals that are collected from 53 ICU patients at a hospital. 
    The patients include 32 females and 21 males with a median age of 64.8.
    PPG signals are recorded using transmission mode devices that are placed on fingertips~\cite{yen2021applying}\cite{schmidt2018introducing}.
    A patient stays still during the collection process with a duration of 8 minutes.
    ECG and PPG signals have a sampling frequency of 125Hz.
 
	\item \textit{CAPNO} \cite{karlen2013CAPNO} is collected from 29 paediatric patients (median age: 8.7) and 13 adult patients (median age: 52.4) during medication examinations at a hospital.
    PPG signals are collected in transmission mode using pulse oximeters placed on fingertips.
    A recording session collects both ECG signals and PPG signals at a sampling frequency of 300Hz.
    The dataset consists of 42 recordings.
    Each recording has a duration of 8 minutes. 
    
	\item \textit{WESAD} \cite{schmidt2018WESAD} is recorded from 15 subjects (3 females, 12 males, mean age: 27.5) for stress and affect detection.
    Subjects perform various activities during the data collection process, \eg{} watching funny videos, delivering speeches and performing arithmetic tasks. 
    A wrist-worn device \textit{Empatica E4} collects PPG and acceleration signals, with a sampling frequency of 64 Hz and 32 Hz, respectively.
    \textit{Empatica E4} collects PPG signals in reflectance mode.
    A chest-worn device records ECG signals with a sampling frequency of 700 Hz.
    The average duration of each subject's recording is 36 minutes.

	\item \textit{DALIA} \cite{Reiss20193079} is collected from 15 subjects (8 females, 7 males, mean age: 30.6)
 while performing eight different daily life activities, \eg{} sitting, walking, cycling, driving and playing table soccer. 
 Similar to collecting the WESAD dataset, PPG and acceleration signals are collected from the wearable device \textit{Empatica E4} in reflectance mode.
 ECG signals are recorded from a chest-worn device.
 PPG, acceleartion and ECG signals are sampled at a frequency of 64Hz, 32Hz, and 700Hz, respectively. 
 Each subject contributes approximately 2 hours of recordings.
 	
\end{itemize}

\vspace{5pt}
\noindent \textbf{Processing PPG Signals.} Since the four datasets are collected with different sampling frequencies, we re-sample PPG signals in the datasets with a sampling frequency of 100Hz, similar to the work by Sarkar and Etemad~\cite{Sarkar_Etemad_2021}.
We apply a 4rd-order Butterworth bandpass filter on PPG signals with a passband frequency of 0.6Hz to 8Hz.
Then, we perform a z-score normalization on the filtered PPG signals.

\begin{figure}[h]
	\begin{subfigure}[t]{0.3\textwidth}
	\includegraphics[width=\linewidth]{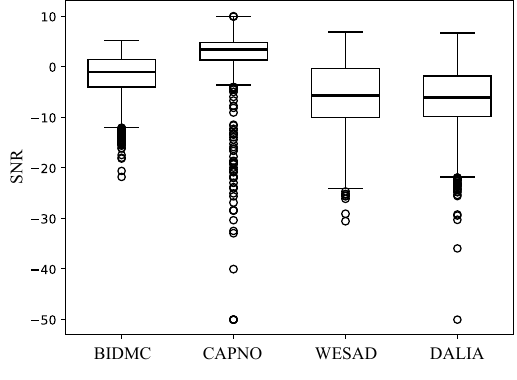}
		\caption{The Four Datasets}
		\label{fig:snr_all}
	\end{subfigure} \hspace{5mm}
    \begin{subfigure}[t]{0.3\textwidth}
		\includegraphics[width=\linewidth]{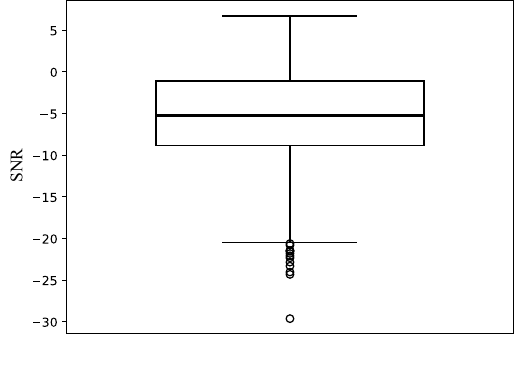}
		\caption{Manually Labeled PPG Segments}
		\label{fig:snr_manual}
	\end{subfigure} \hspace{5mm}
	\begin{subfigure}[t]{0.3\textwidth}
		\includegraphics[width=\linewidth]{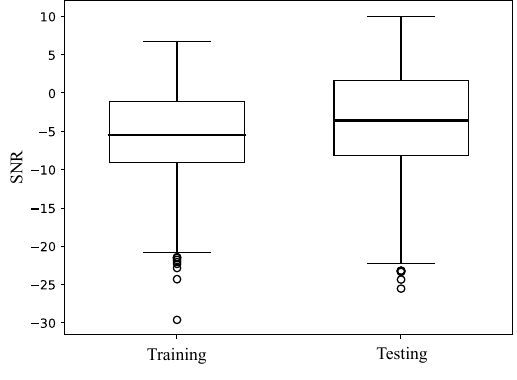}
		\caption{$HR_0$ Dataset}
		\label{fig:snr_label}
	\end{subfigure}
	\caption{Signal-to-noise ratios of our datasets.}
	\label{fig:snr}
	\vspace{-15pt}
\end{figure}

\vspace{5pt}
\noindent \textbf{Signal-to-Noise Ratio (SNR).}
To quantify noise levels in our datasets, we compute SNR using Equation~\ref{eq:snr}~\cite{guo2022effective}.
\begin{equation}\label{eq:snr}
SNR = 10 \, \text{log} \, \frac{P_{Signal}}{P_{Noise}} \, ,
\end{equation}
\noindent where $P_{Signal}$ denotes the power of PPG signals within a range of 5BPM (\ie{} 1/12Hz) centered around the true heart rate, and $P_{Noise}$ represents the power of noise.
A lower SNR value indicates a higher level of noise in PPG signals.
Figure~\ref{fig:snr} shows box plots of SNR values that are calculated using 10-second PPG segments.
Figure~\ref{fig:snr_all} shows SNR values of the four datasets, \ie{} BIDMC, CAPNO, WESAD and DALIA.
Figure~\ref{fig:snr_all} indicates that PPG segments in the datasets WESAD and DALIA have higher levels of noise compared to PPG segments in the datasets BIDMC and CAPNO.

\begin{figure}[h]
	\begin{subfigure}[t]{0.43\textwidth}
		\includegraphics[width=\linewidth]{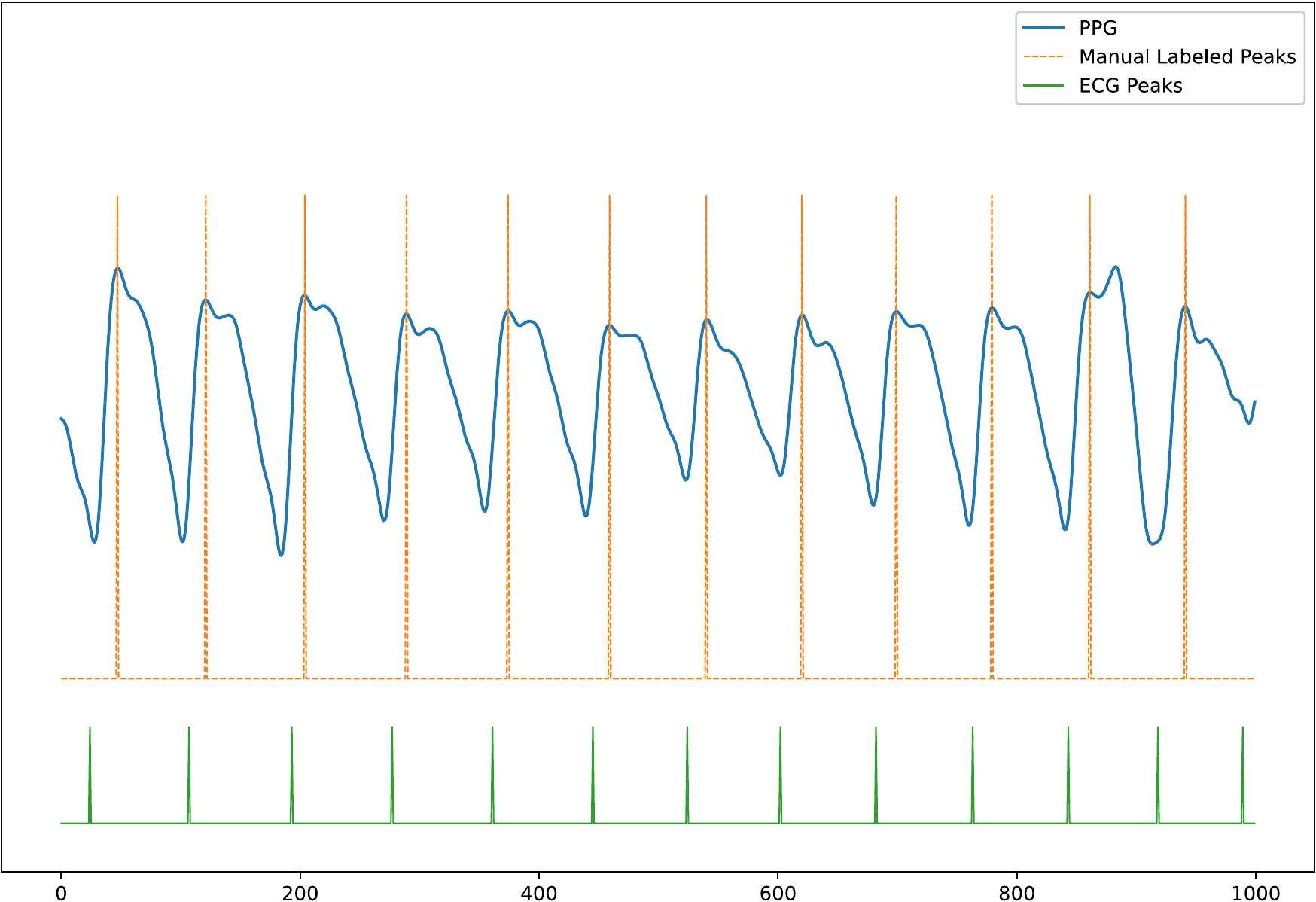}
		\caption{SNR: 2.4}
		\label{}
	\end{subfigure} \hspace{5mm}
	\begin{subfigure}[t]{0.43\textwidth}
		\includegraphics[width=\linewidth]{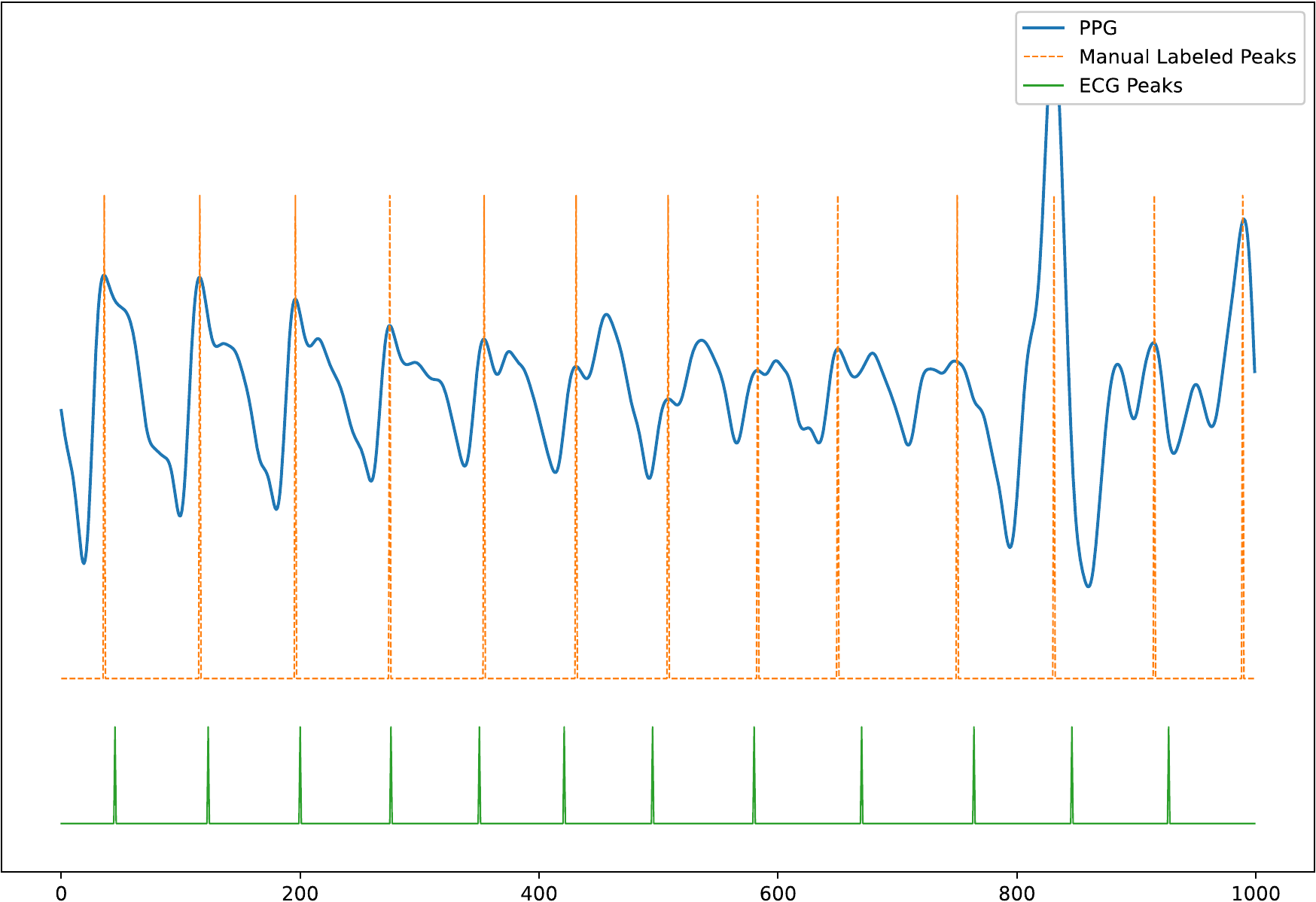}
		\caption{SNR: -3.8}
		\label{}
	\end{subfigure}    
	\medskip
	\begin{subfigure}[t]{0.43\textwidth}
		\includegraphics[width=\linewidth]{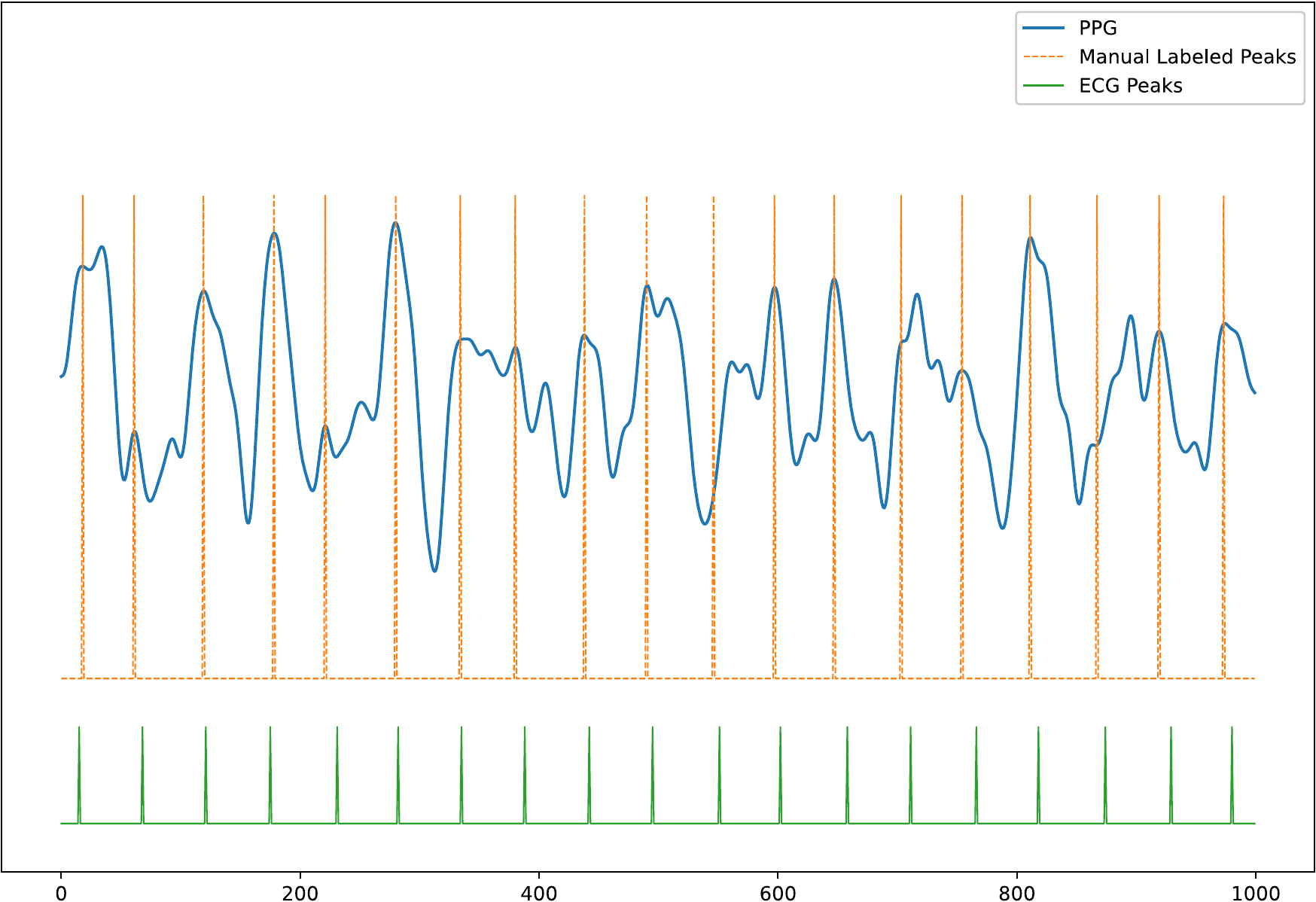}
		\caption{SNR: -6.8}
		\label{}
	\end{subfigure} \hspace{5mm}
	\begin{subfigure}[t]{0.43\textwidth}
		\includegraphics[width=\linewidth]{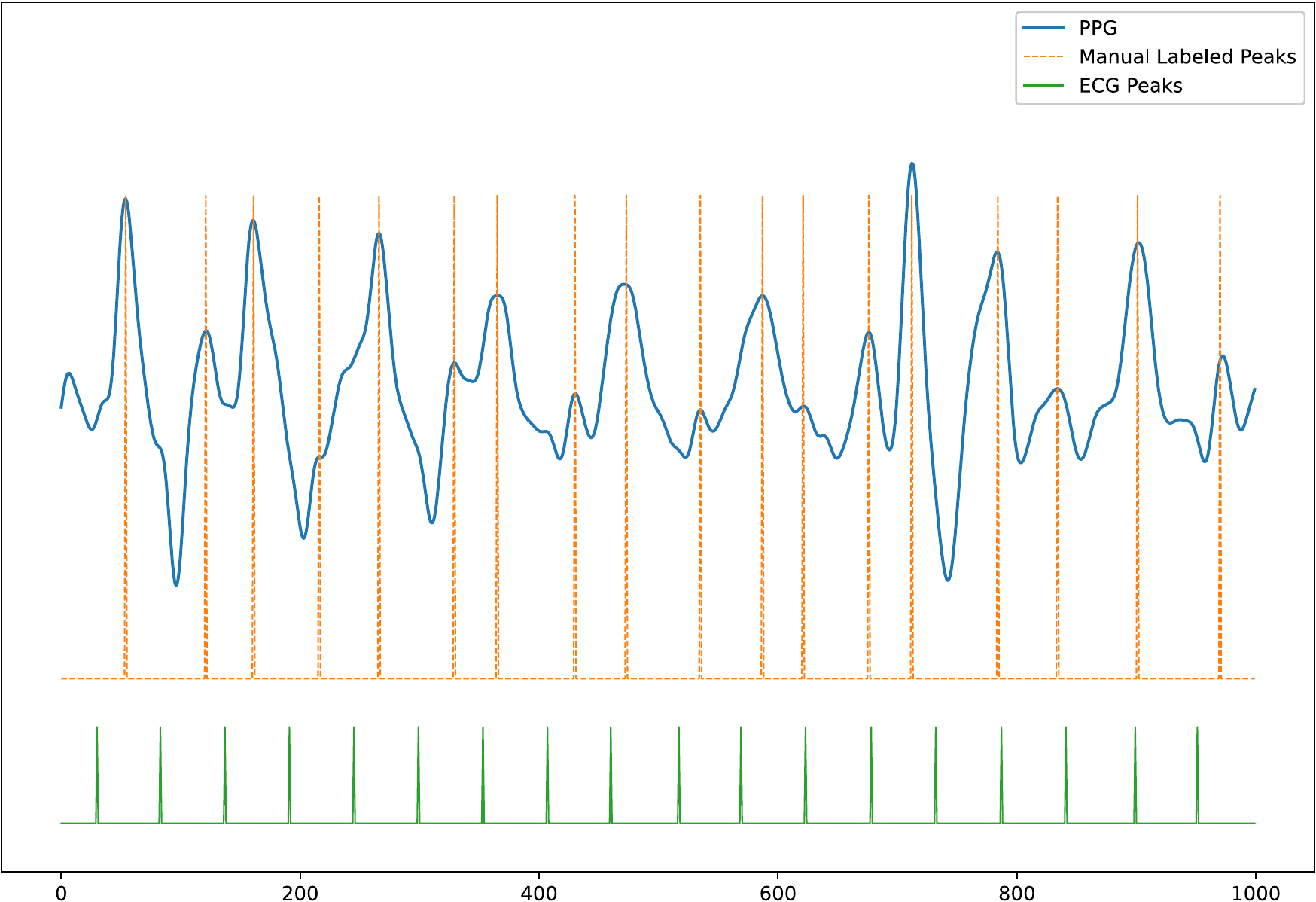}
		\caption{SNR: -12.7}
		\label{fig:peak_hr_4}
	\end{subfigure}
 \vspace{-15pt}
	\caption{Examples of Manually Labeled PPG segments with various SNR values.}
	\label{fig:example_label}
	\vspace{-15pt}
\end{figure}

\vspace{5pt}
\noindent \textbf{Labeling Peaks Manually.}
Due to the precise measurement of heartbeats, ECG signals are commonly used in the literature to provide ground truth heart rates and heart rate variabilities.
Generally, ECG signals and PPG signals are measured from different locations.
For example, in the DALIA dataset, ECG signals are captured from the chest and PPG signals are measured from the wrist using smartwatches.
The varying measurement locations can cause delays in the detection of peaks from ECG signals and PPG signals.
Moreover, existing PPG peak detection approaches cannot accurately identify peaks in motion artifact contaminated PPG signals.
To obtain ground truth peak positions from PPG signals, we use ECG signals as reference signals to manually label peaks~\cite{Kazemi20226054}.
We perform manual labeling on 10-second PPG segments that are randomly sampled from the DALIA dataset.
We choose the DALIA dataset, since DALIA contains clean PPG signals and motion artifact contaminated PPG signals.
Due to high-intensity body movements (\eg{} playing table soccer), motion artifacts may also be observed in ECG signals.
To accurately identify peaks in ECG signals, authors of the DALIA dataset have performed manual inspection to correct peaks in ECG signals~\cite{Reiss20193079}.
We decide PPG peak positions based on amplitude values of PPG signals and intervals between peaks in ECG signals.
If the number of labeled PPG peaks exceeds or falls below 20\% of the number of peaks in ECG signals, we consider our labeling to be incorrect and filter out the PPG segment for labeling.
Note that we have not filtered out any PPG segments in the testing sets for evaluating heart rate estimation and HRV estimation (see Section \textit{Splitting Datasets}).
The first two co-authors independently label peak positions.
Our manual labeling results show that the first two co-authors achieve a Cohen's kappa~\cite{cohen1960coefficient} value of 0.62, indicating a substantial agreement.
The first two co-authors have a discussion session to reach a consensus on the disagreement.
Eventually, we manually labeled a total of 5,145 PPG segments.
Figure~\ref{fig:snr_manual} shows SNR distributions of our manually labeled PPG segments.
The figure illustrates that our labeled PPG segments exhibit a broad range of noise levels, with SNR values ranging from -29.6 to 6.7.
Figure~\ref{fig:snr_label} shows SNR distributions of our labeled PPG segments in the training set and PPG segments in the testing set of the $HR_0$ dataset (described in Section \textit{Splitting Datasets}).
Figure~\ref{fig:snr_label} denotes that our labeled PPG segments in the training set have similar SNR distributions as PPG segments in the testing set.
Figure~\ref{fig:example_label} shows examples of our labeled PPG peaks and ECG peaks in PPG segments with varying SNR values.
Figure~\ref{fig:example_label} demonstrates that PPG peaks may not be synchronized with ECG peaks.
Determining ground truth PPG peaks is challenging since PPG peaks may not always correspond to local maximums due to motion artifacts.
However, we use ECG peaks as a reference to label PPG peaks, thereby ensuring that heart rates derived from PPG peaks and ECG peaks are similar.

\vspace{5pt}
\noindent \textbf{Splitting Datasets.}
To evaluate the effectiveness of our TAU model, we perform the following three tasks.
\begin{itemize}
    \item \textit{Peak detection.} 
We use our manually labeled 5,145 PPG segments as the dataset, \ie{} \textit{peak dataset}, to evaluate the performance of peak detection.
We perform subject-independent experiments~\cite{kwon2019subject} to randomly select 11 subjects from the dataset as the training set, two subjects as the validation set, and the remaining two subjects as the testing set.
The subject-independent experiments ensure that training, validation and testing sets have no PPG segments from the same subject.

\item \textit{Heart rate estimation.}
We combine PPG segments from all the four datasets.
Various studies use different window sizes to estimate heart rates.
To investigate the impact of window size on the performance of our model, we segment PPG signals using two window sizes: 1) 10-second with 1-second overlap~\cite{Sarkar_Etemad_2021}; and 2) 8-second with 6-second overlap~\cite{temko2017accurate}\cite{zhang2015photoplethysmography}\cite{zhang2014troika}.
We perform subject-independent experiments to build two datasets for each window size, \ie{} \textit{$HR_0$ dataset} and \textit{$HR_1$ dataset}.
The training and validation sets of the \textit{$HR_0$ dataset} have the same PPG segments as the \textit{peak dataset}.
To build the training and validation sets with 8-second window size, we segment a labeled 10-second PPG segments into two 8-second PPG segments with 6-second overlap.
To build the \textit{$HR_1$ dataset}, we split subjects into training and testing sets using a ratio of 80:20.
We use 20\% of the training set as the validation set.
The \textit{$HR_0$ dataset} and the \textit{$HR_1$ dataset} have the same testing set.
We ensure that the \textit{$HR_0$ dataset} and the \textit{$HR_1$ dataset} have no common subjects in training, validation and testing sets.
Table~\ref{table:stats} shows the statistics of our datasets.
Our TAU model is trained on the \textit{$HR_0$ dataset} where PPG segments have peak labels.
To obtain a ground truth heart rate label for a PPG segment, we calculate heart rates using ECG signals in the same segment.

\item \textit{HRV estimation.}
We use the $HR_0$ dataset to evaluate the performance of HRV estimation.
Following the previous work~\cite{kwon2018electrocardiogram}\cite{mahdiani201550}, we segment PPG signals from each subject at the testing set into 5-minute lengths with 4-minute overlaps.
We compute reference HRV features from ECG signals in the same time window.

\end{itemize}

\begin{table*}[t]
    \setlength\tabcolsep{1.8pt} 
    \resizebox{0.9\linewidth}{!}{%

	\begin{threeparttable}
	\caption{Statistics of Our Datasets}
	\vspace{-10pt}
	\begin{spacing}{1.0}

        \begin{tabular}{c|c|c|c|c|c|c|c|c}
        \hline
        \textbf{Dataset}               & BIDMC  & CAPNO & WESAD  & DALIA  & \begin{tabular}[c]{@{}c@{}}TR in $HR_0$\end{tabular} & \begin{tabular}[c]{@{}c@{}}TR in $HR_1$\end{tabular} & \begin{tabular}[c]{@{}c@{}}TS \end{tabular} & Total   \\ \hline
        \textbf{\# Subjects}           & 42     & 53    & 15     & 15     & 11                                                   & 100                                                  & 25                                                          & 125     \\ \hline
        \textbf{\# 10-Second PPG Segments} & 2,809  & 2,226 & 9,644  & 14,382 & 4,072                                                &     20,429                                                 &                                                   4,802          & 29,061  \\ \hline
        \textbf{\# 8-Second PPG Segments}  & 12,561 & 9,954 & 43,400 & 64,702 & 8,144                                                & 91,745                                               & 21,560                                                      & 130,617 \\ \hline
        \end{tabular} 
        \end{spacing}
	\label{table:stats}
  	\begin{tablenotes}
		\small
		\item \textit{TR} denotes training sets. \textit{TS} represents testing sets in $HR_0$ and $HR_1$.
	\end{tablenotes}
 \end{threeparttable}
 }
 \vspace{-10pt}
\end{table*}

\subsection{Evaluation Metrics} \label{sec:evaluation_metrics}
\vspace{5pt}
\noindent\textbf{Peak Detection.}
To evaluate the accuracy of peak detection, we use \textit{F1-score}. 
\textit{F1-score} computes the harmonic average of precision and recall.
Precision measures the ratio of correctly detected peaks to the total number of predicted peaks, and recall calculates the fraction of ground truth peaks that a model could identify.
Similar to other works~\cite{elgendi2013systolic}\cite{Reiss20193079}, we employ a tolerance radius to determine whether a predicted peak is correctly identified.
To be specific, if the ground truth peak falls within the tolerance radius around the predicted peak, we denote the predicted peak as a correctly identified peak.
In our experiments, we evaluate $F_1@5$ and $F_1@10$ with a tolerance radius of 5 and 10, respectively.
In PPG segments with a sampling frequency of 100Hz, a tolerance radius of 5 represents a duration of 50ms and a tolerance radius of 10 denotes a duration of 100ms.

\vspace{5pt}
\noindent\textbf{Heart Rate Estimation.}
To evaluate the accuracy of heart rate estimation, we apply the \textit{Mean Absolute Error (MAE)} metric:
\begin{equation}
	MAE = \frac{1}{N}\sum_{i=0}^{N}|\hat{HR}_i-HR_i|,
\end{equation}

\noindent where \textit{N} denotes the total number of PPG segments in the testing set, $\hat{HR}$ represents the estimated heart rate, and $HR$ is the ground truth heart rate.
A lower MAE value indicates better heart rate estimation performance.

\vspace{5pt}
\noindent\textbf{HRV Estimation.}
For each HRV feature (\eg{} the power in the low frequency band), we calculate two metrics:
(1) \textit{Mean Absolute Error} between the HRV feature estimated from PPG signals and the reference HRV feature computed from ECG signals; and (2) Pearson correlation between the estimated HRV feature and the reference HRV feature.
Moreover, we perform Bland-Altman plot analysis~\cite{kwon2018electrocardiogram}\cite{umair2021hrv} to examine the agreement between the two measurements obtained from PPG signals and ECG signals.


\subsection{Hyperparameter Selection}
We optimize hyperparameters of our model using the validation set of the \textit{peak dataset}.
We optimize the loss function in Equation~\ref{eq:smooth_loss} using the optimizer AdamW~\cite{loshchilov2017decoupled}.
We set the initial learning rate of AdamW as 0.002, and the weight decay as 0.05.
The learning rate is constantly adjusted during training using a cosine annealing with warm restarts schedule~\cite{loshchilov2016sgdr}.
We search the number of encoder blocks and decoder blocks $L$ in [1, 2, 3, 4].
We test the threshold value of the peak searching function in [2.5, 5, 7.5, 10]. 
Our TAU model achieves the best validation performance when we set $L=4$ and the threshold value as 7.5.
We empirically set the kernel size of convolution operations as 9 and the number of heads in the multi-head attention mechanism as 2.

\subsection{TAU-lite}
To facilitate the use of our model on devices with limited resources, we develop a lightweight model TAU-lite.  
TAU-lite employs the similar architecture to our TAU model.
Differently, TAU-lite makes the following changes on TAU: 
(1) The number of encoder blocks and decoder blocks $L$ is set as 3; 
(2) We remove the multi-head attention layers in the decoder module; 
(3) A double convolution layer performs one 1D dilated convolution operation; 
(4) We set the number of feature channels $C_0$ as 4 in the encoder module (see Section~\ref{sec:encoder}) and the kernel size of convolution operations as 7; and
(5) We set the embedding size of temporal embeddings as 4 (\ie{} $E=4$).
As a result, the parameter size of TAU-lite reduces to 14.5K.

\subsection{Baselines}

We compare the performance of our model with the following eleven baselines:
\begin{itemize} 
	\item \textbf{Adaptive Threshold}~\cite{kuntamalla2014adaptive} calculates a moving average of the difference between local maxima and local minima, and applies an adaptive threshold filter to select peaks.
 
	\item \textbf{Hilbert}~\cite{chakraborty2020hilbert} applies Hilbert Transform to the second derivative of input PPG segments to identify regions of interest that may contain peaks.
    Peaks are identified as the local maxima in the regions of interest.
 
	\item \textbf{HeartPy}~\cite{VANGENT2019368} calculates a moving average of input PPG segments and uses the moving average as the threshold to determine regions of interest that may contain peaks.
 
	\item \textbf{Elgendi}~\cite{elgendi_norton_brearley_abbott_schuurmans_2013} determines regions of interest by calculating two moving averages with different window sizes. 
    Peaks are identified as maximum values in regions of interest.

    \item \textbf{JOSS}~\cite{zhang2015photoplethysmography} uses sparse signal recovery models to jointly estimate the spectra of PPG signals and acceleration signals. 
    Clean PPG spectra are obtained by subtracting the spectral peaks of acceleration spectra from PPG spectra.
    JOSS tracks spectra peaks in clean PPG spectra to estimate heart rates.
    
    \item \textbf{WFPV}~\cite{temko2017accurate} uses a Wiener filter to obtain clean PPG spectra.
    A phase vocoder and the history of heart rate estimations are used to refine current heart rate estimations.
 
	\item \textbf{1D-CNN}~\cite{Kazemi20226054} applies a stack of 1D dilated convolutions on input PPG segments to predict peak labels.

 \item \textbf{1D-CNN-DT}~\cite{Kazemi20226054} adapts 1D-CNN to use our proposed distance transform labels to supervise the learning of peak label predictions.
 
	\item \textbf{CardioGAN}~\cite{Sarkar_Etemad_2021} is a CycleGAN network that takes PPG segments as inputs to generate ECG segments. The generated ECG segments are used to measure heart rates.
	\item \textbf{DeepPPG}~\cite{Reiss20193079} uses Fast Fourier Transform to convert PPG segments into time-frequency spectra. The converted time-frequency spectra is fed into convolutional neural networks (CNN) to predict heart rates.
	\item \textbf{CorNet}~\cite{8607019cornet} is a deep learning approach that applies two CNN layers and two LSTM layers on input PPG segments to predict heart rates.

\end{itemize}

Our compared baselines fall into three categories: (1) time-domain signal-processing models that perform peak detections, \ie{} Adaptive Threshold, Hilbert, HeartPy, and Elgendi; (2) frequency-domain signal-processing models that estimate heart rates from PPG spectra, \ie{} JOSS and WFPV; and (3) deep learning models, \ie 1D-CNN, 1D-CNN-DT, CardioGAN, DeepPPG, and CorNet.Among the deep learning models, 1D-CNN and 1D-CNN-DT perform peak detection and the other three baselines \ie{} CardioGAN, DeepPPG and CorNet, can only estimate heart rates.
We follow the hyperparameter searching strategies as reported in the papers to search hyperparameters that obtain the best validation performance.

\section{Experimental Results}\label{sec:results}
We perform experiments to answer the following four research questions (RQ).

\vspace{5pt}
\noindent\textbf{RQ1: How does TAU perform on peak detection and heart rate estimation compared to baselines?}

\begin{table}[t]
	\begin{threeparttable}
		\setlength \tabcolsep{5pt} 
			\caption{Peak detection results.}	
                \label{tab:peak_detection}
                \vspace{-10pt}
			\begin{tabular}{c|cccccc}
				\hline
				Model      & F1@5           & Impv.(\%) & F1@10          & Impv.(\%) & MAE           & Impv.(\%) \\ \hline
				Adaptive        & 0.688          & 23.3      & 0.708          & 24.3      & 22.29         & 85.6      \\
				Hilbert         & 0.614          & 38.1      & 0.622          & 41.5      & 23.88         & 86.6      \\
				HeartPy         & 0.735          & 15.4      & 0.754          & 16.8      & 11.17         & 71.4      \\
				Elgendi         & 0.743          & 14.1      & 0.759          & 16.0      & 7.35          & 56.4      \\
				1D-CNN      & 0.835          & 1.5       & 0.854          & 3.1       & 6.54          & 51.1      \\
                    1D-CNN-DT & 0.823 &	3.0 & 0.856 & 2.8 &	5.71 & 43.9 \\
				TAU-lite (ours) & {\ul 0.843}    & 0.6       & {\ul 0.877}    & 0.3       & {\ul 3.83}    & 16.4      \\
				TAU (ours)      & \textbf{0.848} & -         & \textbf{0.880} & -       & \textbf{3.20} & -         \\ \hline
			\end{tabular}
		\begin{tablenotes}
			\vspace{-2pt}
			\small
			\item \textit{Adaptive} denotes \textit{Adaptive Threshold}. \textit{Impv.} denotes the improvement of TAU compared to the baselines. $Impv. = \frac{|metric(B) - metric(TAU)|}{metric(B)}$, where $metric$ represents an evaluation metric, \ie{} F1@5, F1@10 or MAE, and $B$ denotes a baseline. The best results are highlighted in bold and the second best results are underlined (same for other tables).
		\end{tablenotes}
	\end{threeparttable}
	\vspace{-10pt}
\end{table}

\begin{figure}[h]
	\begin{subfigure}[t]{0.49\textwidth}
		\includegraphics[width=\linewidth]{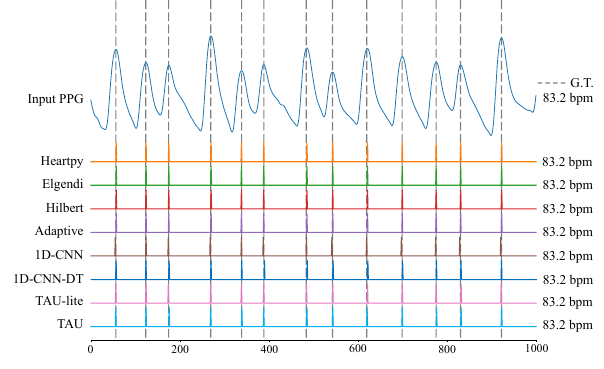}
  
		\caption{BIDMC}
		\label{fig:peak_hr_1}
	\end{subfigure}
	\begin{subfigure}[t]{0.49\textwidth}
		\includegraphics[width=\linewidth]{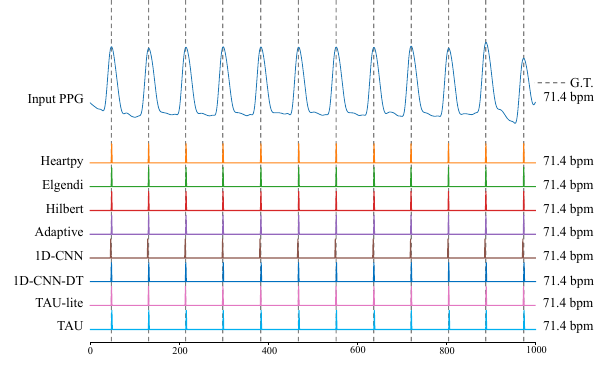}
  
		\caption{CAPNO}
		\label{fig:peak_hr_2}
	\end{subfigure}    
	\medskip
	\begin{subfigure}[t]{0.49\textwidth}
		\includegraphics[width=\linewidth]{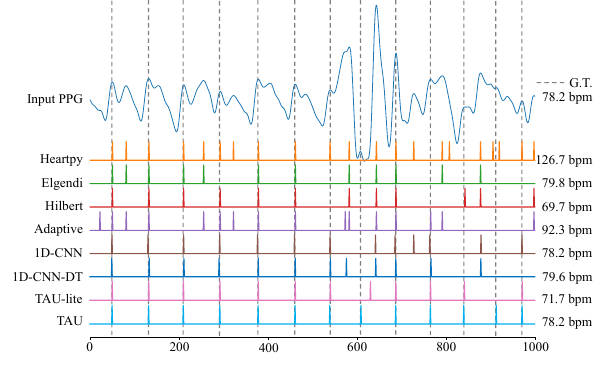}
		\caption{WESAD}
		\label{fig:peak_hr_3}
	\end{subfigure} 
	\begin{subfigure}[t]{0.49\textwidth}
		\includegraphics[width=\linewidth]{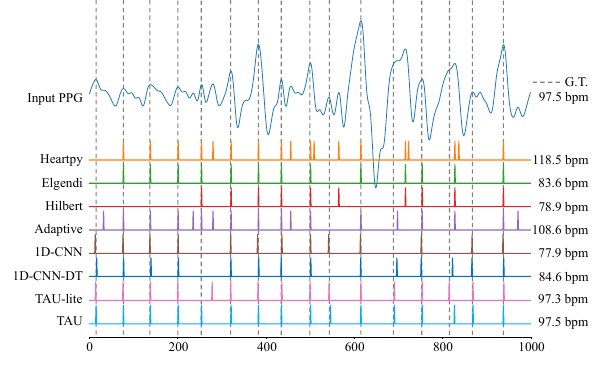}
  
		\caption{DALIA}
		\label{fig:peak_hr_4}
	\end{subfigure}
	\caption{Examples of PPG segments and model prediction results from the testing set of the \textit{$HR_0$ dataset}. PPG segments in the four sub-figures are from the datasets BIDMC, CAPNO, WESAD, and DALIA, respectively. We list estimated heart rates beside peak detection results.
  \textit{Ground Truth (G.T.)} denotes manually labeled peaks and the derived heart rates.}
	\label{fig:peak_hr}
	\vspace{-15pt}
\end{figure}

\vspace{5pt}
\noindent \textbf{Motivation.}
To evaluate the performance, we compare our TAU model with the eleven baselines on two tasks: (1) peak detection  and (2) heart rate estimation.

\vspace{5pt}
\noindent \textbf{Approach.}
We perform the peak detection task using the \textit{peak dataset}.
To evaluate heart rate estimation, we conduct two experiments:
(1) We perform subject-independent experiments on the \textit{$HR_0$ dataset} and \textit{$HR_1$ dataset} using two window sizes, \ie{} 10 seconds and 8 seconds (see Section~\ref{sec:dataset}).
Since the \textit{$HR_1$ dataset} contains ground-truth heart rate labels, CardioGAN, DeepPPG and CorNet are trained on the \textit{$HR_1$ dataset} to optimize heart rate estimations.
(2) We verify the performance of our TAU model on PPG signals with different levels of motion artifacts. 
To quantify the levels of motion artifacts, we equally divide the testing set of the $HR_0$ dataset into four groups based on SNR values, as shown in Figure~\ref{fig:snr_label}.
The four groups represent four different types of motion artifacts. 
Type I, Type II, Type III and Type IV motion artifacts have mean SNR values of 2.8dB (range: 10.1dB to 0.4dB), -2.1dB (range: 0.4dB to -4.6dB), -6.9dB (range: -4.6dB to -9.1dB), and -12.7dB (range: -9.1dB to -50dB), respectively.
We train models using the training set of the $HR_0$ dataset and compare the performance of our model with baselines on each type of motion artifact.
All the above-mentioned experiments are repeated three times. We report the obtained average \textit{F1-score} and \textit{MAE} results.

\vspace{5pt}
\noindent \textbf{Results.}
\textit{Our TAU model consistently outperforms all the eleven baselines on peak detection and heart rate estimation.}

\noindent \textit{1. Peak detection.} Table~\ref{tab:peak_detection} shows the results of our TAU model on the \textit{peak dataset} and the results of baselines that perform peak detection.
From our obtained results, we have the following observations:
(1) Deep learning models (\eg{} 1D-CNN and TAU) outperform signal-processing models (\eg{} HeartPy and Elgendi).
Signal-processing models have no model parameters and find local maxima of PPG signals to determine peaks.
Differently, deep learning models learn patterns of PPG peaks using a substantial number of model parameters;
(2) 1D-CNN-DT achieves better MAE results compared to 1D-CNN, demonstrating the effectiveness of using distance transform labels for peak detection; and
(3) TAU performs better than 1D-CNN and 1D-CNN-DT on peak detection and heart rate estimation.
1D-CNN and 1D-CNN-DT mainly rely on amplitude values to determine peaks.
In addition to the amplitude information, our TAU model models temporal consistency to mitigate large variations in peak-to-peak intervals.
-
Figure~\ref{fig:peak_hr} visualizes our peak detection results on the testing set of the \textit{$HR_0$ dataset}.
In the WESAD example (see Figure~\ref{fig:peak_hr_3}), 1D-CNN incorrectly detects peaks based on the amplitude values of PPG signals between 600$^{th}$ and 800$^{th}$ PPG samples.
Differently, our TAU model is more robust to noise in amplitude values and accurately identifies peak positions.
In the example PPG segment from the DALIA dataset (shown in Figure~\ref{fig:peak_hr_4}), it is observed that 1D-CNN fails to identify three peaks.
Incorporating distance transform labels, 1D-CNN-DT improves the performance and misses identifying one peak.
In contrast, our model could infer peak positions close to the ground truth peaks by leveraging temporal features.

\vspace{5pt}
\noindent \textbf{Discussion on Using Temporal Consistency for PPG Peak Detection.}
To exploit temporal consistency, many existing works (\eg{} 1D-CNN) detect peaks in the peak detection phase and then employ the peak correction phase that incorporates heuristic approaches to filter out invalid peaks~\cite{Kazemi20226054}\cite{VANGENT2019368}\cite{xu2019photoplethysmography}\cite{zhang2014troika}.
The approaches may use threshold values to constrain maximum and minimum distances between two successive peaks and avoid large peak-to-peak variations.
Despite effective for specific scenarios, these heuristic approaches mainly have the following two limitations:
(1) The peak correction phase may require a dozen threshold values to be manually specified~\cite{chung2018finite}\cite{temko2017accurate}\cite{zhang2015photoplethysmography}.
The optimal choice of threshold values may vary from datasets, which poses challenges for the approaches to adapt to real-life scenarios where users may perform various daily physical activities; and
(2) Since the peak correction phase usually occurs after peaks are detected, retained peaks that are determined by amplitude values cannot be ensured to be accurate.
Moreover, previous estimation errors may propagate and affect current peak corrections~\cite{chung2018finite};
Unlike existing approaches that apply temporal consistency during the peak correction phase, we model temporal consistency in the peak detection phase that uses the time module and distance transform labels to assist in peak detection.
As shown in Figure~\ref{fig:peak_hr_3} and Figure~\ref{fig:peak_hr_4}, our model effectively combines global temporal information and contextual amplitude values to accurately infer peak positions.

\begin{table}[t]
    \setlength\tabcolsep{1.8pt} 
    \resizebox{\linewidth}{!}{%
	\begin{threeparttable}
		\begin{spacing}{1.1}
			\caption{MAE comparison between TAU and the baselines on heart rate estimations.}
			\label{tab:heart_rate_estimation}
                \begin{tabular}{c|c|cccccc|cccccc}
                \hline
                \multirow{2}{*}{Model} & \multirow{2}{*}{D.S.}                        & \multicolumn{6}{c|}{10-Second PPG Segments}                                                           & \multicolumn{6}{c}{8-Second PPG Segments}                                                 \\ \cline{3-14} 
                                       &                                              & BIDMC         & CAPNO         & WESAD         & DALIA         & Avg.          & Impv.(\%)             & BIDMC         & CAPNO         & WESAD         & DALIA         & Avg.          & Impv.(\%) \\ \hline
                Adaptive               & \multirow{13}{*}{$HR_0$}                     & 2.63          & 8.51          & 25.70         & 22.34         & 14.80         & 84.7                  & 3.26          & 8.68          & 25.44         & 22.44         & 14.95         & 81.5      \\
                Hilbert                &                                              & 2.34          & 0.61          & 18.67         & 24.79         & 11.60         & 80.5                  & 3.04          & 0.77          & 17.92         & 23.95         & 11.42         & 75.7      \\
                HeartPy                &                                              & 1.85          & 1.05          & 12.82         & 11.55         & 6.82          & 66.9                  & 2.53          & 1.23          & 13.35         & 12.08         & 7.30          & 62.1      \\
                Elgendi                &                                              & \textbf{1.25} & {\ul 0.39}    & 6.22          & 8.40          & 4.06          & 44.3                  & {\ul 1.75}    & 0.36          & 6.64          & 8.85          & 4.40          & 37.0      \\
                JOSS                   &                                              & 2.38          & 0.40          & 10.12         & 11.59         & 6.12          & 63.1                  & 10.90         & \textbf{0.32} & 7.93          & 9.95          & 7.27          & 61.9      \\
                WFPV                   &                                              & 2.30          & 1.63          & 6.06          & 9.00          & 4.75          & 52.4                  & 2.85          & 0.39          & 6.02          & 8.28          & 4.38          & 36.8      \\
                1D-CNN                 &                                              & {\ul 1.27}    & 0.46          & 6.71          & 7.15          & 3.90          & 42.1                  & 1.77          & 0.43          & 6.68          & 7.37          & 4.06          & 31.8      \\
                1D-CNN-DT  &   &    1.63	& 0.78	& 5.82	& 6.00	& 3.56	& 36.5                                                   & \textbf{1.69} & 0.61          & 5.72          & 6.66          & 3.67          & 24.5      \\
                CardioGAN              &                                              & 1.76          & 3.40          & 4.57          & 4.36          & 3.52          & 35.8                  & 1.94          & 2.33          & 5.13          & {\ul 4.87}    & 3.57          & 22.4      \\
                DeepPPG                &                                              & 2.31          & 1.85          & 5.67          & 6.22          & 4.01          & 43.6                  & 2.53          & 0.64          & 5.53          & 5.92          & 3.66          & 24.3      \\
                CorNet                 &                                              & 1.96          & 3.49          & {\ul 3.80}    & {\ul 4.38}    & 3.41          & 33.7                  & 2.92          & 1.79          & 5.17          & 5.48          & 3.84          & 27.9      \\
                TAU-lite (ours)        &                                              & 1.33          & 1.20          & 4.15          & 4.69          & {\ul 2.84}    & 20.4                  & 1.82          & 0.52          & {\ul 4.85}    & 4.93          & {\ul 3.03}    & 8.6       \\
                TAU (ours)             &                                              & 1.30          & \textbf{0.36} & \textbf{3.54} & \textbf{3.84} & \textbf{2.26} & \textbf{-}            & 1.90          & {\ul 0.34}    & \textbf{4.48} & \textbf{4.36} & \textbf{2.77} & -         \\ \hline
                CardioGAN              & \multicolumn{1}{l|}{\multirow{3}{*}{$HR_1$}} & 1.57          & 0.87          & 4.64          & 5.00          & 3.02          & -                     & 1.88          & 0.71          & 4.41          & 5.90          & 3.22          & -         \\
                DeepPPG                & \multicolumn{1}{l|}{}                        & 2.01          & 1.05          & 4.70          & 5.65          & 3.35          & -                     & 2.15          & 0.98          & 4.76          & 5.68          & 3.39          & -         \\
                CorNet                 & \multicolumn{1}{l|}{}                        & 1.70          & 0.83          & 3.01          & 3.84          & 2.34          & -                     & 2.31          & 1.14          & 3.79          & 4.08          & 2.83          & -         \\ \hline
                \end{tabular}
        \end{spacing}
	\begin{tablenotes}
        \vspace{-2pt}
        \small
        \item\textit{D.S.} represents datasets. \textit{Avg.} denotes the average MAE results that are obtained from the testing sets of BIDMC, CAPNO, WESAD, and DALIA. The best results and the second best results obtained from the $HR_0$ dataset are highlighted.
        \textit{Impv.} denotes the average MAE improvement of our TAU model over the baselines, \ie{}
        $Impv. = \frac{Avg.(B) - Avg.(TAU)}{Avg.(B)}$.
	\end{tablenotes}
\end{threeparttable}
}
\vspace{-10pt}
\end{table}

\vspace{5pt}
\noindent \textit{2. Heart rate estimation.}
Table~\ref{tab:heart_rate_estimation} shows the performance comparison between TAU and the eleven baselines on heart rate estimation.
We discuss the detailed findings in the following.

\begin{itemize} 
	\item \textit{Baselines that predict heart rates (\eg{} CorNet) can obtain better heart rate estimation performance than baselines that perform peak detection (\eg{} 1D-CNN).}
	CorNet can estimate heart rates from the frequency domain without identifying each peak in the time domain. 
	In contrast, peak detection models (\eg{} 1D-CNN) may produce inaccurate peak positions and therefore generate imprecise heart rate estimations.
	
    \item \textit{CardioGAN, DeepPPG and CorNet achieve better heart rate estimation results on the \textit{$HR_1$ dataset} compared to the \textit{$HR_0$ dataset}.}
    In general, more training data leads to better generalization on unseen data~\cite{zhao2022modeling}.
	Moreover, we observe that on the dataset with less motion artifacts (\ie{} CAPNO dataset), CorNet performs worse (MAE=0.83) than the signal-processing model Elgendi (MAE=0.39). 
	For example, on a clean PPG segment as shown in Figure~\ref{fig:peak_hr_2}, CorNet estimates a heart rate as 69.5 bpm with an absolute error of 1.9.
	In addition to the frequency of peaks, the other frequencies in a PPG segment may affect model estimations.

    \item \textit{Deep learning models generally achieve better heart rate estimation performance using 10-second PPG segments compared to 8-second PPG segments.}
    A 10-second PPG segment contains more contextual information, which is helpful for deep learning models to estimate heart rates.
    Different from deep learning models, frequency-domain signal-processing models JOSS and WFPV achieve better MAE on datasets CAPNO, WESAD, and DALIA using 8-second PPG segments.
    PPG segments with a duration of 8 seconds have 6-second overlaps (see Section~\ref{sec:dataset}).
    The large overlaps are beneficial for the tracking modules in JOSS and WFPV to correct heart rate estimations using the history of heart rate estimations.
    However, previous estimation errors may propagate to current heart rate estimations~\cite{chung2018finite}, as Table~\ref{tab:heart_rate_estimation} shows that JOSS achieves the worst performance (MAE = 10.9) at the BIDMC dataset;

	\item \textit{Our TAU model achieves the best heart rate estimation performance using both 10-second PPG segments and 8-second PPG segments.}
        On datasets with 10-second PPG segments, TAU improves the best performing baseline \ie{} CorNet, by an average margin of 33.7\%.
        On datasets with 8-second PPG segments, TAU outperforms the best performing baseline, CardioGAN, by an average margin of 22.4\%.
        TAU outperforms all the other baselines on WESAD and DALIA datasets that are collected from reflectance mode devices.
        TAU that is trained on the \textit{$HR_0$ dataset} achieves similar performance as CorNet that is trained with five times (MAE=2.34) and eleven times (MAE=2.83) the amount of data from the \textit{$HR_1$ dataset}.  
        Moreover, we observe that our lightweight model TAU-lite achieves the second best performance on the \textit{$HR_0$ dataset} for both window sizes. 
        Since we observe similar performance using datasets with the two window sizes (\ie{} 10 seconds and 8 seconds), in the following RQs, we present our results obtained on the $HR_0$ dataset with 10-second PPG segments.
\end{itemize}

\begin{table}[t]
    \setlength\tabcolsep{1.8pt} 
    \resizebox{0.65\linewidth}{!}{%

	\begin{threeparttable}
		\begin{spacing}{1.1}
			\caption{Heart rate estimation results (MAE) on the four types of motion artifacts.}
			\label{tab:ma_types}
                \begin{tabular}{c|cc|cc|cc|cc}
                \hline
                \multirow{2}{*}{Model} & \multicolumn{2}{c|}{Type I} & \multicolumn{2}{c|}{Type II} & \multicolumn{2}{c|}{Type III} & \multicolumn{2}{c}{Type IV} \\ \cline{2-9} 
                                       & Avg.             & Impv.(\%)   & Avg.             & Impv.(\%)    & Avg.              & Impv.(\%)    & Avg.             & Impv.(\%)   \\ \hline
                Adaptive               & 17.57            & 96.4        & 19.03            & 90.6         & 21.56             & 80.7         & 22.78            & 63.5        \\
                Hilbert                & 8.80             & 92.9        & 15.94            & 88.8         & 20.57             & 79.8         & 28.49            & 70.8        \\
                HeartPy                & 1.02             & 38.3        & 4.96             & 64.0         & 10.87             & 61.7         & 16.36            & 49.1        \\
                Elgendi                & 1.03             & 39.1        & 4.71             & 62.0         & 7.61              & 45.3         & 13.32            & 37.5        \\
                JOSS              & 0.79 &	20.1&	3.14&	43.1	&10.89&	61.8&	25.29&	67.1  \\
                WFPV              & 3.12 &	79.8 &	4.02 &	55.5 &	5.95 &	30.0 &	14.16 &	41.2  \\
                1D-CNN                 & 1.02             & 38.4        & 3.59             & 50.2         & 6.52              & 36.1         & 10.73            & 22.4        \\
                1D-CNN-DT              & 1.01             & 37.9        & 3.79             & 52.8         & 6.73              & 38.1         & 10.04            & 17.1        \\
                CardioGAN              & 2.02             & 68.9        & 2.29             & 22.0         & 4.55              & 8.5          & {\ul 8.50}       & 2.1         \\
                DeepPPG                & 1.35             & 53.3        & 2.34             & 23.5         & 5.44              & 23.4         & 13.02            & 36.1        \\
                CorNet                 & 2.08             & 69.8        & {\ul 1.89}       & 5.6          & \textbf{3.98}     & -4.6         & 9.04             & 7.9         \\
                TAU-lite (ours)        & {\ul 0.67}       & 6.7         & 1.99             & 10.1         & 4.56              & 8.6          & 9.38             & 11.3        \\
                TAU (ours)             & \textbf{0.63}    & -           & \textbf{1.79}    & -            & {\ul 4.16}        & -            & \textbf{8.32}    & -           \\ \hline
                \end{tabular}
        \end{spacing}
	\begin{tablenotes}
        \vspace{-2pt}
        \small
        \item 
	\end{tablenotes}
\end{threeparttable}
}
\vspace{-10pt}
\end{table}

\noindent \textbf{Results on Four Types of Motion Artifacts.}
Table~\ref{tab:ma_types} shows MAE comparison between TAU and the baselines on the four types of motion artifacts using the $HR_0$ dataset with 10-second PPG segments.
We observe that as the noise level increases from Type I motion artifacts (mean SNR: 2.8dB) to Type IV motion artifacts (mean SNR: -12.7dB), models achieve worse performance on heart rate estimations.
Our TAU model achieves the best heart rate estimation performance in both low-noise-level signals (\ie{} SNRs higher than -4.6dB) and high-noise-level signals (\ie{} SNRs lower than -9.1dB).
On PPG signals with Type III motion artifacts, TAU outperforms all models except CorNet.
Our results demonstrate that TAU is robust in heart rate estimation across varying noise levels.

\vspace{5pt}
\noindent \textbf{Why TAU is Robust to Motion Artifacts.}
We attribute the improvement of our model to the deployment of the following three components:
(1) the encoder-decoder architecture similar to U-Net.
The encoder module encodes contextual amplitude information through convolutional layers, and the decoder module captures the global context of PPG signals to refine peak label predictions using the attention mechanism.
Our model captures both local and global features of PPG signals;
(2) distance transform labels. Unlike hard labels with binary values to denote non-peak labels and peak labels, distance transform labels consist of multiple integer label values that form a wave with an oscillating shape (see Figure~\ref{fig:Distance Transform}).
Therefore, predicting a distance transform label requires a model to learn temporal and amplitude dependencies among contextual PPG signals;
and (3) the time module. Identifying peaks based on amplitude values in noisy PPG signals is prone to errors. 
To mitigate the errors, we incorporate a time module in our architecture to enforce temporal consistency, ensuring that neighboring peak-to-peak intervals have no large variations.

\begin{table}[t]
	\setlength\tabcolsep{1.8pt} 
	\vspace{5pt}
	\begin{spacing}{1.3}
		\caption{HRV Comparison between TAU and the baselines.}
  	\label{tab:hrv}
            \tiny
		\resizebox{\textwidth}{!}{%
              \begin{tabular}{c|c|cc|cc|cc|cc|cc|cc|cc}
                \hline
                \multirow{2}{*}{Dataset} & \multirow{2}{*}{Model} & \multicolumn{2}{c|}{Mean NN} & \multicolumn{2}{c|}{RMSSD} & \multicolumn{2}{c|}{SDNN} & \multicolumn{2}{c|}{SDSD} & \multicolumn{2}{c|}{LF} & \multicolumn{2}{c|}{HF} & \multicolumn{2}{c}{LF/HF} \\ \cline{3-16} 
                 &  & MAE & R & MAE & R & MAE & R & MAE & R & MAE & R & MAE & R & MAE & R \\ \hline
                \multirow{6}{*}{BIDMC} & HeartPy & 1.11 & 0.990 & 11.39 & 0.603 & 8.15 & 0.590 & 11.05 & - & 107.92 & - & 242.53 & - & 0.58 & 0.463 \\
                 & Elgendi & 3.22 & 0.996 & 15.96 & 0.940 & 10.93 & 0.926 & 14.50 & 0.763 & 5.55 & 0.692 & 15.17 & 0.788 & 0.73 & - \\
                 & 1D-CNN & 0.72 & {\ul 0.997} & 4.86 & 0.902 & 3.36 & 0.882 & 4.49 & 0.792 & \textbf{3.23} & \textbf{0.651} & \textbf{9.03} & \textbf{0.646} & \textbf{0.37} & \textbf{0.929} \\
                 & 1D-CNN-DT & \textbf{0.14} & \textbf{0.999} & {\ul 2.15} & {\ul 0.952} & \textbf{1.33} & 0.948 & {\ul 1.75} & {\ul 0.913} & 3.05 & 0.593 & 4.47 & 0.878 & 0.41 & 0.878 \\
                 & TAU-lite (ours) & {\ul 0.20} & \textbf{0.999} & \textbf{1.85} & \textbf{0.954} & {\ul 1.14} & \textbf{0.953} & \textbf{1.41} & \textbf{0.926} & 2.39 & 0.620 & 4.73 & 0.894 & 0.49 & 0.838 \\
                 & TAU (ours) & 0.37 & {\ul 0.997} & 2.52 & 0.912 & 1.43 & 0.925 & 1.79 & 0.903 & {\ul 2.41} & {\ul 0.641} & {\ul 4.85} & {\ul 0.683} & {\ul 0.40} & {\ul 0.901} \\ \hline
                \multirow{6}{*}{CAPNO} & HeartPy & 1.07 & 0.997 & 12.33 & - & 8.28 & 0.598 & 12.12 & - & 10.30 & 0.624 & 22.49 & - & 1.27 & - \\
                 & Elgendi & 3.16 & {\ul 0.999} & 20.52 & - & 13.37 & - & 19.56 & - & 5.59 & - & 6.58 & - & 1.41 & - \\
                 & 1D-CNN & 0.70 & {\ul 0.999} & 9.08 & - & 5.73 & - & 9.06 & - & 12.48 & - & 10.68 & - & 1.36 & - \\
                 & 1D-CNN-DT & 0.22 & {\ul 0.999} & 4.06 & - & 2.56 & - & 3.88 & - & 5.28 & - & 3.16 & - & {\ul 0.75} & \textbf{0.839} \\
                 & TAU-lite (ours) & {\ul 0.11} & \textbf{1.000} & {\ul 2.41} & - & {\ul 1.42} & - & {\ul 2.20} & - & {\ul 1.01} & {\ul 0.753} & {\ul 1.00} & - & \textbf{0.74} & {\ul 0.790} \\
                 & TAU (ours) & \textbf{0.08} & \textbf{1.000} & \textbf{2.12} & - & \textbf{1.28} & - & \textbf{1.98} & - & \textbf{0.63} & \textbf{0.855} & \textbf{0.94} & - & 0.79 & 0.773 \\ \hline
                \multirow{6}{*}{WESAD} & HeartPy & 4.61 & 0.800 & 34.20 & 0.321 & 21.42 & 0.297 & 28.13 & 0.251 & 2980.79 & - & 6602.92 & - & 0.75 & - \\
                 & Elgendi & 5.89 & 0.643 & 32.54 & - & 19.61 & - & 24.23 & - & 1624.27 & - & 2976.44 & - & 0.75 & - \\
                 & 1D-CNN & 4.25 & 0.694 & 26.92 & 0.364 & 16.01 & {\ul 0.388} & 21.40 & {\ul 0.272} & 149.06 & - & 312.55 & - & 0.76 & - \\
                 & 1D-CNN-DT & 2.84 & 0.749 & 22.53 & 0.268 & 13.39 & 0.217 & 18.09 & 0.180 & 1287.80 & - & 2372.93 & - & {\ul 0.73} & - \\
                 & TAU-lite (ours) & \textbf{1.77} & \textbf{0.920} & {\ul 12.35} & {\ul 0.393} & {\ul 6.91} & 0.331 & {\ul 9.12} & 0.219 & {\ul 40.36} & - & {\ul 85.80} & - & \textbf{0.66} & {\ul 0.290} \\
                 & TAU (ours) & {\ul 1.87} & {\ul 0.913} & \textbf{9.51} & \textbf{0.499} & \textbf{5.12} & \textbf{0.577} & \textbf{6.58} & \textbf{0.448} & \textbf{16.24} & \textbf{0.264} & \textbf{28.44} & - & \textbf{0.66} & \textbf{0.365} \\ \hline
                \multirow{6}{*}{DALIA} & HeartPy & 4.85 & 0.922 & 25.51 & - & 15.57 & - & 19.68 & - & 335.28 & - & 754.32 & - & 2.37 & 0.271 \\
                 & Elgendi & 6.70 & 0.915 & 25.74 & - & 15.31 & - & 17.49 & - & 48.92 & - & 96.55 & - & 2.44 & - \\
                 & 1D-CNN & 3.98 & 0.947 & 20.46 & - & 12.14 & - & 15.54 & - & 46.04 & - & 89.90 & - & 2.41 & 0.163 \\
                 & 1D-CNN-DT & 3.38 & 0.952 & 17.18 & - & 10.16 & - & 13.06 & - & 45.17 & - & 88.95 & - & 2.36 & {\ul 0.390} \\
                 & TAU-lite (ours) & {\ul 2.10} & {\ul 0.971} & {\ul 12.01} & - & {\ul 6.73} & - & {\ul 9.28} & - & {\ul 25.32} & {\ul 0.227} & {\ul 47.12} & - & \textbf{2.08} & \textbf{0.423} \\
                 & TAU (ours) & \textbf{1.86} & \textbf{0.978} & \textbf{10.61} & - & \textbf{5.89} & - & \textbf{8.35} & - & \textbf{19.88} & \textbf{0.319} & \textbf{33.08} & - & {\ul 2.13} & 0.367 \\ \hline
                \end{tabular}
               
		}
  \end{spacing}
  	\begin{threeparttable}
		\begin{tablenotes}
			\small
			\item $R$ denotes Pearson correlation coefficients. Statistically significant $R$ values are presented ($p < 0.01$).  "-" represents that $R$ values are not statistically significant.
		\end{tablenotes}
	\end{threeparttable}
	\vspace{5pt}
\end{table}

\vspace{5pt}
\noindent \textbf{RQ2: How does TAU perform on HRV estimation compared to baselines?}

\vspace{5pt}
\noindent \textbf{Motivation.}
To evaluate the performance on HRV estimation, we compare our TAU model with the best four baselines that perform peak detections, \ie{} HeartPy, Elgendi, 1D-CNN and 1D-CNN-DT.

\vspace{5pt}
\noindent \textbf{Approach.}
We train deep learning models on the training set of the $HR_0$ dataset.
To compute peak-to-peak intervals in a 5-minute PPG segment, we detect peaks in each 10-second PPG segment.
From the obtained peak-to-peak intervals, we derive time-domain HRV features and frequency-domain HRV features, similar to the work by Kuang~\etal~\cite{kuang2023shuffle}.
In particular, time-domain HRV features include mean peak-to-peak intervals (\ie{} Mean NN), root mean squared difference of successive peak-to-peak intervals (\ie{} RMSSD), standard deviation of peak-to-peak intervals (\ie{} SDNN), standard deviation of peak-to-peak interval differences (\ie{} SDSD).
Frequency-domain HRV features comprise the power in the low-frequency band (\ie{} LF), the power in the high-frequency band (\ie{} HF), and the ratio of LF to HF (\ie{} LF/HF).

\vspace{5pt}
\noindent \textbf{Results.}
\textit{Our TAU model can accurately estimate HRV from PPG signals with low levels of noise.}
Table~\ref{tab:hrv} presents the MAE results and Pearson correlation coefficients (\ie{} R values) obtained from each model on the testing sets of the four datasets (\ie{} BIDMC, CAPNO, WESAD, and DALIA).
In Appendix~\ref{app:bland}, we show Bland-Altman plot analysis \cite{bland1986statistical} for our TAU model.
We observe that models achieve consistent Pearson correlation results in the four datasets. 
For example, all the compared models obtain significant Pearson correlation coefficients on HRV features in the BIDMC dataset.
In particular, 1D-CNN-DT, TAU-lite and TAU models achieve R values higher than 0.9 for Mean NN, RMSSD, SDNN and SDSD features.
However, in both the CAPNO and DALIA datasets, Pearson correlation coefficients for RMSSD, SDNN, SDSD, and HF features are statistically insignificant across all the models.
We manually examine PPG signals in the CAPNO dataset, and find that the dataset contains many outliers with high levels of noise (shown in Figure~\ref{fig:snr_all}).
The outliers greatly impact the accuracy in estimating HRV features from PPG signals.
In the DALIA dataset, subjects perform various exercises to collect PPG signals.
Estimating HRV features from PPG signals with high levels of motion artifacts is challenging~\cite{georgiou2018can}.
Moreover, our TAU model achieves the best MAE results for both time-domain measurements and frequency-domain measurements in the CAPNO, WESAD and DALIA datasets.
TAU-lite achieves the best Pearson correlation results among the compared models for time-domain HRV measurements in the BIDMC dataset.
Our results show that our model improves the accuracy in estimating HRV features from PPG signals.

\vspace{5pt}
\noindent \textbf{RQ3: Is the use of the attention mechanism, the distance transform, and the time module beneficial to TAU?}

\vspace{5pt}
\noindent \textbf{Motivation.}
To verify the impact of the attention mechanism, the distance transform, and the time module on the performance of our TAU model, we perform an ablation study.

\vspace{5pt}
\noindent \textbf{Approach.}
To study the effectiveness of the components in our TAU model, we compare the performance of our model and three variants on peak detection and heart rate estimation using the \textit{peak dataset} and \textit{$HR_0$ dataset}.
We describe the three variants in the following.

\begin{itemize}
	\item $TAU_{b}$ is the base version of our TAU model. $TAU_{b}$ makes the following architecture changes on TAU: (1) $TAU_{b}$ removes the time module in our TAU model (see Section~\ref{sec:time}). The decoder module takes the outputs $\bm{x_e^L}$ from the encoder module as inputs to generate peak label predictions; 
    (2) $TAU_{b}$ removes multi-head attention layers (see Section~\ref{sec:decoder}) in the decoder module; and
    (3) $TAU_{b}$ is trained using labels that are generated with the hard labeling schema (see Section~\ref{sec:dt}).
	
	\item \textit{$TAU_b$+Att} modifies the variant $TAU_{b}$ to add multi-head attention layers in the decoder module.
	
	\item \textit{$TAU_b$+Att+DT} modifies the variant \textit{$TAU_b$+Att} to use distance transform labels to supervise model training.
	
\end{itemize}

\noindent \textbf{Results.}
\textit{the attention mechanism, the distance transform, and the time module are beneficial to our TAU model.}
Table~\ref{tab:ablation_study} shows F1-score of peak detection and MAE results of heart rate estimation.
From our obtained results, we have the following observations:
(1) Among the compared variant models, F1-score has no significant difference. 
We conjecture that in noisy PPG signals, variant models may not identify peaks within the tolerance radius of manually labeled peaks.
Nevertheless, we observe that our TAU model has significant MAE improvements over the three variant models;
(2) \textit{$TAU_b$+Att} achieves better MAE results than $TAU_{b}$ on the \textit{peak} dataset and the \textit{$HR_0$} dataset.
The attention mechanism updates embeddings of a PPG sample by aggregating embeddings of all PPG samples.
The improvement shows that the attention mechanism can better capture global dependencies among PPG samples;
(3) Compared to \textit{$TAU_b$+Att}, using distance transform labels in \textit{$TAU_b$+Att+DT} can further improve the performance of estimating heart rates.
Variant models \textit{$TAU_b$+Att} and \textit{$TAU_b$+Att+DT} have the same number of parameters with the only difference of ground truth labels.
Without bringing extra computation cost to estimate heart rates, \textit{$TAU_b$+Att+DT} outperforms \textit{$TAU_b$+Att} by an average margin of 41.4\% and 30.3\% on the \textit{peak} dataset and \textit{$HR_0$} dataset, respectively.
The continuity of distance transform labels allows a smooth transition from peak samples to non-peak samples. 
Therefore, semantically similar PPG samples can be assigned with similar predicted labels; and
(4) Finally, incorporating the time module in our TAU model achieves the best MAE values.
The time module learns to improve representations of PPG samples with temporal embeddings to capture the temporal consistency of PPG signals.
Therefore, the time module can be viewed as a regularizer to regularize the variations among peak-to-peak intervals.

\begin{table}[t]
	\setlength\tabcolsep{1.8pt} 
	\vspace{-5pt}
		\caption{ Comparison between TAU and the three variants on peak detection and heart rate estimations.}
		\resizebox{\linewidth}{!}{%
			\begin{tabular}{c|cccccc|cccccc}
				\hline
				& \multicolumn{6}{c|}{Peak Dataset} & \multicolumn{6}{c}{$HR_0$ Dataset (MAE) } \\
				\multirow{-2}{*}{Model} & {\color[HTML]{333333} F1@5} & Impv. (\%) & {\color[HTML]{333333} F1@10} & Impv. (\%) & MAE & Impv. (\%) & BIDMC & CAPNO & WESAD & DALIA & Avg. & Impv. (\%) \\ \hline
				$TAU_{b}$ & 0.851 & -0.3 & 0.873 & 0.8 & 6.31 & 49.3 & 1.75 & 0.46 & 6.13 & 7.08 & 3.85 & 41.4 \\
				$TAU_b$+Att & \textbf{0.853} & -0.6 & 0.877 & 0.4 & 6.02 & 46.8 & 1.80 & 0.43 & 5.28 & 6.37 & 3.47 & 34.8 \\
				{$TAU_b$+Att+DT} & {\ul 0.850} & -0.2 & \textbf{0.881} & -0.1 & {\ul 3.53} & 9.2 & {\ul 1.64} & \textbf{0.32} & {\ul 3.76} & {\ul 3.96} & {\ul 2.42} & 6.6 \\
				TAU (ours) & 0.848 & - & {\ul 0.880} & - & \textbf{3.20} & - & \textbf{1.30} & {\ul 0.36} & \textbf{3.54} & \textbf{3.84} & \textbf{2.26} & - \\ \hline
			\end{tabular}
		}
  	\begin{threeparttable}
		\begin{tablenotes}
			\vspace{-2pt}
			\small
			\item \textit{on $HR_0$ dataset, we show MAE results of heart rate estimations.}
		\end{tablenotes}
	\end{threeparttable}
		\label{tab:ablation_study}
	\vspace{-10pt}
\end{table}

\vspace{5pt}
\noindent \textbf{RQ4. How our TAU model assigns attention weights on PPG samples and learns to predict peak labels?}

\vspace{5pt}
\noindent \textbf{Motivation.}
We use the attention mechanism in the time module and decoder module to capture dependencies among PPG samples.
Moreover, our TAU model predicts peak labels at the encoder module, the time module and the decoder module.
To understand how our model captures sample dependencies and learns to predict peak labels, we perform a qualitative analysis.

\vspace{5pt}
\noindent \textbf{Approach.}
We take an example PPG segment from the testing set of the \textit{peak dataset} to visualize attention weights and model outputs.
We obtain attention weights from the first decoder block.
Since we use two heads in the multi-head attention mechanism, we visualize attention weights that are averaged over the two heads.
We resize the shape of attention weights to $1000 \times 1000$ for ease of visualization, where 1000 is the number of PPG samples in a PPG segment.

\begin{figure}
	\centering
	\includegraphics[width=0.7\linewidth]{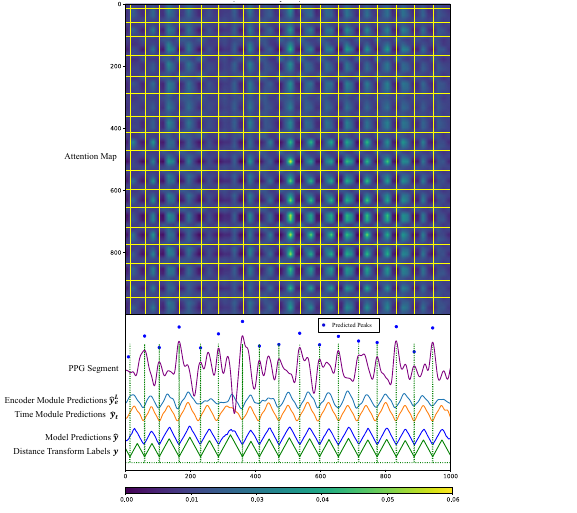}
	
	\caption{Visualization of our obtained attention weights and model outputs.} 
	\label{fig:attention}
\end{figure}

\vspace{5pt}
\noindent \textbf{Results.}
\textit{1. Non-peak PPG samples pay high attention to other non-peak samples.}
Figure~\ref{fig:attention} uses an attention map to illustrate the distribution of the obtained attention weights on PPG samples.
In the attention map, a pixel at the $i^{th}$ row and $j^{th}$ column represents an assigned attention weight between a query PPG sample $i$ and a key PPG sample $j$.
Attention weights at each row sum to one.
The pixel color represents the value of attention weights.
Brighter colors indicate larger attention weights and darker colors indicate smaller attention weights.
We use yellow lines to denote that the PPG samples are ground truth peaks.
From the attention map in Figure~\ref{fig:attention}, we have the following observations:
(1) Non-peak samples tend to attend to other non-peak samples.
For example, many PPG samples pay attention to PPG samples at around 500$^{th}$ column, as many areas at the 500$^{th}$ column have bright colors;
and (2) Peak samples tend to not assign high attention values on non-peak samples.
For example, at the 500$^{th}$ column, PPG samples at bright areas may not include peak samples.
Our results show that TAU may project the embeddings of peak samples and non-peak samples to distinct semantic space.

\vspace{5pt}
\noindent \textit{2. Label predictions are evolved to get close to ground truth labels.}
Figure~\ref{fig:attention} shows that we cannot clearly identify peaks from the predictions $\bm{\hat{y}_e^L}$ given by the encoder module.
The predictions are improved in the time module, as the label predictions $\bm{\hat{y}_t}$ from the time module can clearly indicate peak positions.
However, the time module performs predictions solely based on temporal features without the amplitude information of PPG signals (see Section~\ref{sec:time}). 
Adding the amplitude information in the decoder module corrects label predictions.
We observe that peaks that are extracted from model predictions $\bm{\hat{y}}$ closely match with ground truth peaks.

\vspace{5pt}
\noindent \textbf{RQ5. How efficient is TAU compared to baselines?}

\vspace{5pt}
\noindent \textbf{Motivation.}
To evaluate the efficiency of TAU, we compare the inference time between our TAU model and the eleven baselines on the \textit{peak dataset}.

\noindent \textbf{Approach.}
On the testing set of the \textit{peak dataset}, we compute the average inference time on a PPG segment with a window size of 10 seconds.
We evaluate the inference time of our TAU model and the eleven baselines on a machine with Intel Xeon CPUs and Nvidia Tesla V100 GPUs.
Signal-processing models, \ie{} Adaptive Threshold, Hilbert, HeartPy, Elgendi, JOSS and WFPV perform peak detection on CPUs.
The other deep learning models perform inference on GPUs.

\begin{table}[t]
    \setlength\tabcolsep{1.8pt} 
		\caption{Inference time comparison between TAU and the baselines.}
		\label{tab:efficiency}
		\vspace{-5pt}
		\resizebox{\linewidth}{!}{%
                \begin{tabular}{c|ccccccccccccc}
                \hline
                Model     & Adaptive & Hilbert & HeartPy & Elgendi & JOSS & WFPV & 1D-CNN & \multicolumn{1}{c}{\begin{tabular}[c]{@{}c@{}}1D-CNN\\ -DT\end{tabular}} & CardioGAN & DeepPPG & CorNet & \begin{tabular}[c]{@{}c@{}}TAU-lite\\ (ours)\end{tabular} & \begin{tabular}[c]{@{}c@{}}TAU\\ (ours)\end{tabular} \\ \hline
                \# Params & -        & -       & -       & -       & -    & -    & 3.2K   & 3.2K                                                                         & 28.3M     & 2.2M    & 262K   & 14.5K                                                     & 2.2M                                                 \\
                IT (ms)   & 2.1      & 2.6     & 3.7     & 0.8     & 4192.0      & 0.2     & 3.9    & 3.9                                                                         & 25.4      & 6.2     & 19.7   & 7.3                                                       & 20.8                                                 \\ \hline
                \end{tabular}		
		}
	\begin{threeparttable}
		\begin{tablenotes}
			\vspace{-2pt}
			\small
			\item \textit{\# Params} denotes the number of parameters in a model and \textit{IT} indicates the inference time.
		\end{tablenotes}
	\end{threeparttable}
	\vspace{-10pt}
\end{table}

\vspace{5pt}
\noindent \textbf{Results.}
\textit{TAU-lite has similar efficiency compared to the baseline 1D-CNN.}
Table~\ref{tab:efficiency} shows the average inference time of a 10-second PPG segment.
From our obtained results, we have the following findings.
\begin{itemize} 
\item \textit{Among the eleven baselines, the signal-processing baselines are the most efficient baselines.} 
Signal-processing models are efficient since the models have no model parameters.
However, JOSS takes the longest time to perform inference, since JOSS requires intensive computation on large matrices~\cite{biswas2019heart}.

\item \textit{Our TAU model achieves a similar inference time to CorNet and CardioGAN.}
Each of the three models contains components that may incur extra inference time. 
Specifically, our TAU model adopts the multi-head attention mechanism, which is computationally more intensive than convolution operations.
CorNet has LSTM layers that require sequential processing of sample features.
CardioGAN uses 13 times more parameters (\ie{} 28.3M) than our TAU model (\ie{} 2.2M) to generate ECG signals. 

\item \textit{Our TAU-lite model has similar inference time compared to 1D-CNN and DeepPPG.}
TAU-lite, 1D-CNN and DeepPPG obtain an average inference time of 7.3ms, 3.9ms, and 6.2ms, respectively.
Compared to 1D-CNN, our TAU-lite model requires an additional 3.4ms of inference time.
However, TAU-lite outperforms 1D-CNN by 41.6\% on heart rate estimation (see Table~\ref{tab:peak_detection}).
\end{itemize}

\section{Conclusion}\label{sec:conclusion}
In this paper, we propose the TAU model to perform peak detection from PPG signals.
Our TAU model adopts an encoder-decoder based architecture.
We leverage the temporal consistency of PPG signals to design a time module that augments encoder outputs with temporal features.
Temporal features help our model generalize to unseen PPG segments when amplitude values are contaminated by noises.
In the decoder module, we use the multi-head attention mechanism to capture global dependencies among PPG samples.
To endow semantic similar PPG samples with similar embeddings, we use distance transform labels to supervise the learning of peak predictions.
We perform subject independent experiments to evaluate the performance of our TAU model.
Our results show that TAU outperforms all the eleven baselines on peak detection and heart rate estimation.
Specifically, on datasets with 10-second PPG segments, our TAU model improves the best performing baseline \ie{} CorNet, by an average margin of 33.7\%.
With 20\% of training data, TAU achieves similar heart rate estimation performance as the CorNet model.
Our experiments on HRV analysis demonstrate that our model can accurately estimate HRV features from PPG signals with low levels of noise.
Moreover, our results show that our time module augments the amplitude information of PPG signals to improve the performance of heart rate estimation.

Our datasets are collected from 125 subjects while conducting a variety of daily life activities, such as perfoming physical exercises and solving arithmetic tasks.
The experimental results obtained from these datasets demonstrate the usability of our model in real-world settings.
In the future, we are interested to deploy our models on wearable devices to study the real-time performance.

\bibliographystyle{ACM-Reference-Format}
\bibliography{sample-base}

\appendix

\section{Bland-Altman Plot Analysis}\label{app:bland}
\begin{figure}[h]
	\begin{subfigure}[t]{0.24\textwidth}
		\includegraphics[width=\linewidth]{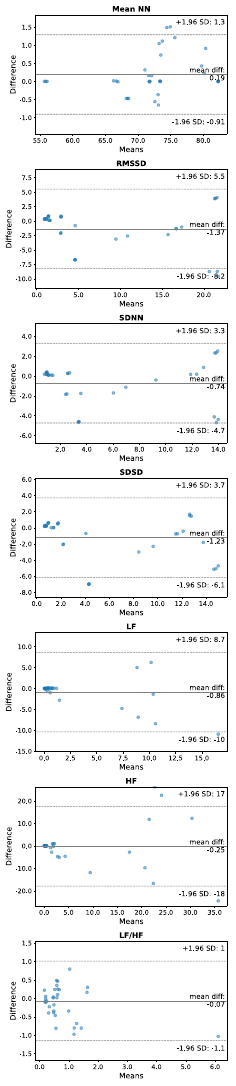}
		\caption{BIDMC}
		\label{fig:bland-altman-bidmc}
	\end{subfigure}
	\begin{subfigure}[t]{0.24\textwidth}
		\includegraphics[width=\linewidth]{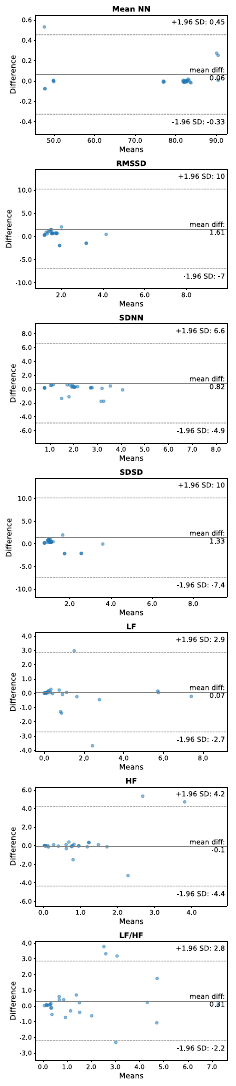}
		\caption{CAPNO}
		\label{fig:bland-altman-capno}
	\end{subfigure}    
	\begin{subfigure}[t]{0.24\textwidth}
		\includegraphics[width=\linewidth]{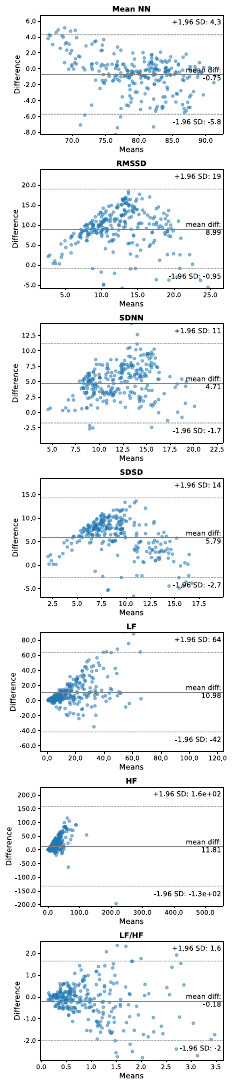}
		\caption{WESAD}
		\label{fig:bland-altman-wesad}
	\end{subfigure} 
	\begin{subfigure}[t]{0.24\textwidth}
		\includegraphics[width=\linewidth]{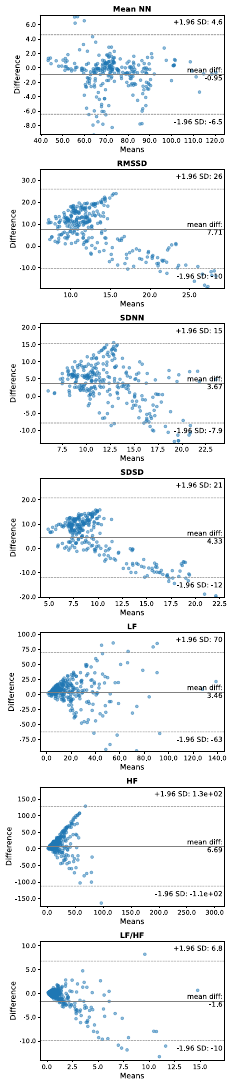}
		\caption{DALIA}
		\label{fig:bland-altman-dalia}
	\end{subfigure}
	\caption{Bland-Altman Plots of our TAU model for HRV Analysis}
	\label{fig:Bland-Altman}
	\vspace{-15pt}
\end{figure}

Figure \ref{fig:Bland-Altman} shows the Bland-Altman plots \cite{bland1986statistical} of HRV features that are obtained from our TAU model on the four datasets \ie{} BIDMC, CAPNO, WESAD, and DALIA. 
A Bland-Altman plot shows the difference between the HRV feature that is measured from PPG signals and the reference HRV feature that is measured from ECG signals.
The x-axis represents the average of the two measurements and the y-axis shows the difference between the two measurements.
A Bland-Altman plot determines the limits of agreement, representing a range within which 95\% of differences between the two measurements are observed.


From our obtained results, we observe that on BIDMC and CAPNO datasets, HRV features estimated by our TAU model have strong agreement with reference HRV features that are measured by ECG signals.
For example, the mean differences between the two measurements in all HRV features are close to zero.
On the WESAD and DALIA datasets, the HRV feature \textit{Mean NN} obtains low mean differences of -0.75 and -0.95, respectively, suggesting an agreement between the two measurements.
However, on other HRV features, the mean differences are high and limits of agreements display a wide range.
Our results show that accurately estimating HRV features from noisy PPG signals is challenging.


\end{document}